\documentclass[lettersize,journal]{IEEEtran}
\usepackage{amsmath,amsfonts}
\usepackage{algorithmic}
\usepackage{algorithm}
\usepackage{array}
\usepackage[caption=false,font=normalsize,labelfont=sf,textfont=sf]{subfig}
\usepackage{textcomp}
\usepackage{stfloats}
\usepackage{url}
\usepackage{verbatim}
\usepackage{graphicx}
\usepackage{cite}
\usepackage{amsfonts}
\usepackage{booktabs}
\usepackage{multirow}
\usepackage{makecell}
\usepackage{color}
\usepackage{forest}
\usetikzlibrary{arrows.meta,shapes,positioning,shadows,trees}
\usepackage{lscape}
\usepackage{orcidlink}
\usepackage{hyperref}
\usepackage{marvosym}
\usepackage{graphicx}
\usepackage[capitalise]{cleveref}
\usepackage{enumitem}
\usepackage{amssymb}
\usepackage{amsthm}
\usepackage{amsmath}
\usepackage{xspace}
\newtheorem{defi}{Definition}
\DeclareMathOperator*{\argmax}{arg\,max}

\newcommand{\transitionspace}{\mathcal{P}}

\newcommand{\obsspace}{\mathcal{O}}
\newcommand{\obsdensity}{\Omega}
\newcommand{\statespace}{\mathcal{S}}
\newcommand{\actionspace}{\mathcal{A}}

\newcommand{\rewardfunc}{r}

\newcommand{\alg}{\texttt{ED2}\xspace}

\begin{document}

\title{ED2: Environment Dynamics Decomposition \\World Models for Continuous Control}

\author{
Jianye Hao\,\orcidlink{0000-0002-0422-8235},~\IEEEmembership{Senior Member,~IEEE}, Yifu Yuan\,\orcidlink{0009-0009-2194-942X}, Cong Wang\,\orcidlink{0000-0001-8501-5778}, Zhen Wang\,\orcidlink{0000-0002-8182-2852},~\IEEEmembership{Fellow,~IEEE}
        % <-this % stops a space
\thanks{Manuscript received February 14, 2024.}
\thanks{Corresponding authors: Jianye Hao (\href{mailto:jianye.hao@tju.edu.cn}{jianye.hao@tju.edu.cn}), Zhen Wang (\href{mailto:w-zhen@nwpu.edu.cn}{w-zhen@nwpu.edu.cn})}
\thanks{
Jianye Hao, Yifu Yuan, Cong Wang are with the College of Intelligence and Computing, Tianjin University, Tianjin 300350, China (e-mail: 
 \href{mailto:jianye.hao@tju.edu.cn}{jianye.hao@tju.edu.cn};
 \href{mailto:yuanyf@tju.edu.cn}{yuanyf@tju.edu.cn};
 \href{mailto:wangcong@tju.edu.cn}{wangcong@tju.edu.cn}).

Zhen Wang is with the School of Artificial Intelligence, OPtics and Electronics
(iOPEN) \& School of Cyberspace, Northwestern Polytechnical University, Xi’an 710072, China (e-mail: \href{mailto:w-zhen@nwpu.edu.cn}{w-zhen@nwpu.edu.cn}).
}

} % <-this % stops a space
% Montreal Institute of Learning Algorithms (MILA), Quebec, Canada
 % \thanks{Manuscript received January 19, 2024}}

        % <-this % stops a space

\markboth{Journal of \LaTeX\ Class Files,~Vol.~14, No.~8, August~2021}%
{Shell \MakeLowercase{\textit{et al.}}: A Sample Article Using IEEEtran.cls for IEEE Journals}


\maketitle

\begin{abstract}

Model-based reinforcement learning (MBRL)  achieves significant sample efficiency in practice in comparison to model-free RL, but its performance is often limited by the existence of model prediction error. 
To reduce the model error, standard MBRL approaches train a single well-designed network to fit the entire environment dynamics, but this wastes rich information on multiple sub-dynamics which can be modeled separately, allowing us to construct the world model more accurately. 
In this paper, we propose \textbf{E}nvironment \textbf{D}ynamics \textbf{D}ecomposition~(\alg), a novel world model construction framework that models the environment in a decomposing manner. 
\alg contains two key components: \textit{sub-dynamics discovery} (SD2) and \textit{dynamics decomposition prediction} (D2P). SD2 discovers the sub-dynamics in an environment automatically and then D2P constructs the decomposed world model following the sub-dynamics. \alg can be easily combined with existing MBRL algorithms and empirical results show that \alg significantly reduces the model error, increases the sample efficiency, and achieves higher asymptotic performance when combined with the state-of-the-art MBRL algorithms on various continuous control tasks. Our code is open source and available at \url{https://github.com/ED2-source-code/ED2}.
\end{abstract}


\section{Introduction} \label{section:introduction}
\IEEEPARstart{R}{einforcement} Learning (RL) is a general learning framework for solving sequential decision-making problems and has made significant progress in many fields \cite{DBLP:journals/nature/MnihKSRVBGRFOPB15,DBLP:journals/nature/SilverHMGSDSAPL16,DBLP:journals/nature/VinyalsBCMDCCPE19,DBLP:journals/corr/abs-1911-08265}. In general, RL methods can be divided into two categories regarding whether a world model is constructed for the policy deriving: model-free RL (MFRL) and model-based RL (MBRL).
MFRL methods train the policy by directly interacting with the environment, resulting in low sample efficiency. On the contrary, MBRL methods learn a dynamics model of environments from experience. The policy uses the dynamic model to act as a simulator, generating experience for optimization.
Through imagination, we can identify appropriate actions with low trial-and-error costs~\cite{luo2024survey}. MBRL produces impressive sample efficiency by modeling the environment, but often with limited asymptotic performance and suffers from the model error~\cite{DBLP:conf/icml/Lai00020, DBLP:conf/iclr/KaiserBMOCCEFKL20}.

Existing MBRL algorithms can be divided into four categories according to the paradigm they follow:
the first category focuses on generating imaginary data by the world model and training the policy with these data via MFRL algorithms \cite{DBLP:conf/nips/KidambiRNJ20, DBLP:conf/nips/YuTYEZLFM20, DBLP:conf/icml/WangWJH23}; the second category leverages the differentiability of the world model, and generates differentiable trajectories for policy optimization \cite{DBLP:conf/icml/DeisenrothR11, DBLP:conf/icml/LevineK13, DBLP:conf/nips/ZhuZLZ20}; the third category aims to obtain an accurate value function by generating imaginations for temporal difference (TD) target calculation \cite{DBLP:conf/nips/BuckmanHTBL18, DBLP:journals/corr/abs-1803-00101}; the last category of works focuses on reducing the computational cost of the policy deriving by combining the optimal control algorithm (e.g. model predictive control) with the learned world models \cite{DBLP:conf/nips/ChuaCML18, DBLP:conf/corl/OkadaT19, DBLP:journals/corr/abs-2008-05556, hansen2022temporal}.
Regardless of paradigms, most MBRL methods use the learned dynamic model to generate substantial samples to guide policy learning. Therefore, it is critical to learn a dynamics model that accurately simulates the underlying dynamics of the real environment. The more accurate the world model is, the more reliable data can be generated, and finally, the better policy performance can be achieved.

\begin{figure*}[t]
\centering
\includegraphics[width=1\linewidth]{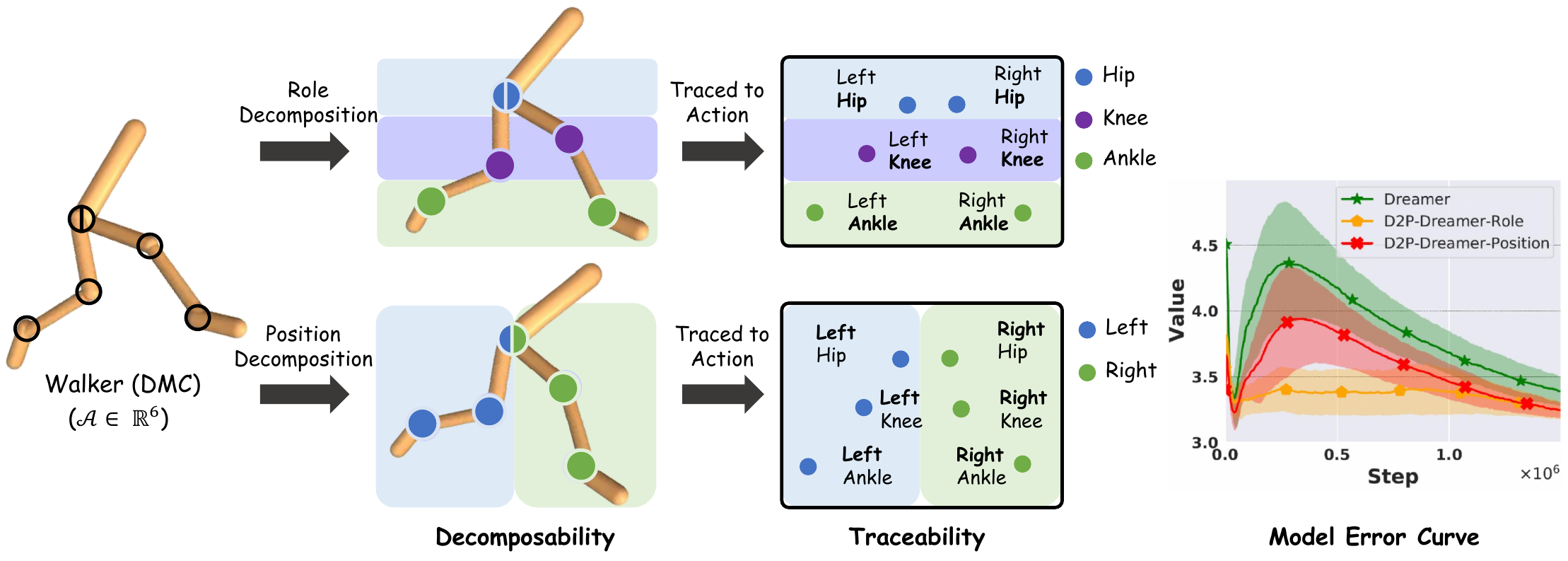}
\caption{\textbf{Motivation Example.} The Walker task has six action dimensions, including $\{\textit{hip}, \textit{knee}, \textit{ankle}\}$ for the $\{\textit{left}, \textit{right}\}$ limbs. \textbf{\textit{Decomposability}}: The dynamics can be decomposed into multiple sub-dynamics in various ways, each sub-dynamics is described with different background colors. Each circle corresponds to a joint including single or multiple action dimensions. \textbf{\textit{Traceability}}: Dynamics can be traced to the impact caused by the action, and for each sub-dynamics, we show the meanings of the action dimensions it traced to. \textbf{\textit{Model Error Curve}}:
The model error comparison on the Walker task of original Dreamer and D2P-Dreamer methods with different decomposition show that modeling each decomposed sub-dynamics separately can significantly
reduce the model error.
}
\label{fig:motivation_example}
\end{figure*}

To this end, various advanced architectures have been proposed to improve the model prediction accuracy. For example, rather than directly predict the next state, some works construct a world model for the state change prediction \cite{DBLP:conf/iclr/LuoXLTDM19, DBLP:conf/iclr/KurutachCDTA18}. Model ensemble is also widely used in model construction for uncertainty estimation, which provides a more reliable prediction \cite{DBLP:conf/nips/JannerFZL19, DBLP:conf/nips/PanHT020}. To reduce the model error in long trajectory generation, optimizing multi-step prediction errors~\cite{DBLP:conf/icml/HafnerLFVHLD19, DBLP:journals/corr/abs-1905-13320} or building jumpy prediction~\cite{zhang2023leveraging} are also effective techniques.
However, these techniques improve the environment modeling in a black-box way, which ignores the inner decomposed structure of environment dynamics. According to kinesiology studies~\cite{d2003combinations, ting2007neuromechanics, ficuciello2016synergy}, the human central nervous system (CNS) activates muscles in groups, extracting muscle synergies and reducing the complexity of controlling muscles separately. Similar to muscle synergy, we can decompose the dynamics in groups to simplify the complexity of control and dynamics modeling. For example, \cref{fig:motivation_example} shows the Walker robot from DMC Suite tasks~\cite{DBLP:journals/corr/abs-1801-00690}, where the dynamics can be decomposed in various ways. We can intuitively split complex dynamics into various decompositions: according to the role of sub-dynamics, we can decompose it into: $\{\textit{hip}, \textit{knee}, \textit{ankle}\}$; alternatively, according to the position of sub-dynamics, we can decompose it into: $\{\textit{left}, \textit{right}\}$. The model error curve shows that no matter whether we decompose the Walker task according to role or position, modeling each decomposed sub-dynamics separately can significantly reduce the model error of the existing MBRL algorithm.

Inspired by the above example, we propose environment dynamics decomposition (\alg), a novel world model construction framework that models the dynamics in a decomposing fashion. ED2 contains two main components: \textit{sub-dynamics discovery} (SD2) and \textit{dynamics decomposition prediction} (D2P).
SD2 is proposed to decompose the dynamics into multiple sub-dynamics, which can be flexibly designed and we also provide three alternative approaches: complete decomposition, human prior, and the clustering-based method.
D2P is proposed to construct the world model from the decomposed dynamics in SD2, which models each sub-dynamics separately in an end-to-end training manner.
\alg is orthogonal to existing MBRL algorithms and can be used as a backbone to easily combine with any MBRL algorithm. We combine the ED2 framework with various types of MBRL algorithm~\cite{DBLP:conf/nips/JannerFZL19, hansen2022temporal, DBLP:conf/iclr/HafnerLB020, hafner2020mastering} in both image- and state-based environments. Experiments show that \alg improves the accuracy of the model and increases the performance significantly when combined with existing MBRL algorithms.

\section{Related Work}
\label{section:related_work}

Model based RL reduces the sample complexity of the learning process by building the environmental dynamics model as a simulator~\cite{luo2024survey}. Then the dynamics model can be used for planning~\cite{ebert2018visual, schrittwieser2020mastering, yu2020mopo, hafner2019learning, hafner2019dream}, accelerating model-free policy learning using generated data~\cite{pong2018temporal, DBLP:conf/nips/JannerFZL19, ha2018recurrent} or more accurate estimation of value functions~\cite{feinberg2018model, buckman2018sample}. However, model errors tend to weaken the performance of MBRL approaches. Even if the prediction is accurate for a single step, when the rollout length increases, the model error will compound and accumulate to move out of the region of high accuracy predictions~\cite{lai2020bidirectional}. Inaccurate model predictions can cause disruptions to the learning process, leading to a collapse in performance. 

As a result, a lot of work~\cite{DBLP:conf/nips/ChuaCML18, thananjeyan2020safety, lai2020bidirectional, nair2020goal} has studied how to mitigate the model's prediction error. A mainstream manner is by measuring and utilizing the uncertainty of the dynamics model. Ensemble models~\cite{DBLP:conf/nips/ChuaCML18} are able to effectively prevent the predictive inaccuracy of any single model through uncertainty estimation. \cite{kalweit2017uncertainty} uses generated data only when the critic has high uncertainty. Some exploration methods~\cite{kalweit2017uncertainty, sekar2020planning} can also be used to collect the data where the model prediction is uncertain. Furthermore, M2AC~\cite{pan2020trust} measures uncertainty in dynamics and discards forecasts with high uncertainty. Some methods have been proposed to mitigate the compound error problem. 
\cite{asadi2019combating} and \cite{whitney2018understanding} use a multi-step dynamics model to directly predict the results after the impact of multiple actions in a state. Similarly, the segment-based model~\cite{mishra2017prediction} makes predictions for the entire trajectory. WGAN ~\cite{wu2019model} and DWM~\cite{diffuionworldmodel} attempt to learn the model by matching the multi-step rollout distribution instead of step-by-step autoregressive prediction. Besides, The MBPO~\cite{DBLP:conf/nips/JannerFZL19} avoided the compounding error by generating short branched rollouts from real states where it has accurate predictions. Following the MBPO, BMPO~\cite{lai2020bidirectional} proposes a bi-directional prediction model to utilize both the forward model and backward model for short branched generation.
These approaches are limited to investigate how to better model the dynamics in a black-box manner, but they all ignore the nature of the environmental dynamics themselves. \alg stands on the novel perspective of environmental dynamics and kinematics to reduce the model prediction error, which is orthogonal to the previous methods and can be well combined with various MBRL algorithms.

\section{Preliminary}
\subsection{Reinforcement Learning}
Given an environment, we can define a finite-horizon partially observable Markov decision process (POMDP) as $(\statespace, \actionspace, \rewardfunc, \transitionspace, \gamma, \obsspace, \obsdensity, T)$, where $\statespace \in \mathbb{R}^n$ is the state space and $\actionspace \in \mathbb{R}^m$ is the action space, $\rewardfunc : S \times A \rightarrow \mathbb{R}$ denotes the reward function,
$\transitionspace : S \times A \rightarrow S$ denotes the environment dynamics, $\gamma$ is the discount factor.
The agent receives an observation $o \in \Omega$, which contains partial information about the state $s \in \statespace$.
$\obsspace$ is the observation function, which mapping states to probability distributions over observations. The decision process length is denoted as $T$. Let $\mathbb{\eta}$ denote the expected return of a policy $\pi$ over the initial state distribution $\rho_0$.
The goal of an RL agent is to find the optimal policy $\pi^*$ which maximizes the expected return:
\begin{equation*}
    \pi^* = \mathop{\arg\max}_{\pi}\mathbb{\eta}[\pi] = \mathop{\arg\max}_{\pi}\mathbb{E}_\pi[\sum_{t=0}^{T}\gamma^tR(s_t, a_t)],
\end{equation*}
where $s_0 \sim \rho_{0}$, $o_t \sim \obsspace(\cdot|s_t)$, $a_t \sim \pi(\cdot|o_t)$,  $s_{t+1} \sim P(\cdot|s_t, a_t)$.
If the environment is fully observable, i.e., $\Omega = \statespace$ and $\obsspace$ is an identity function, POMDP is equivalent to the MDP as $(\statespace, \actionspace, \rewardfunc, \transitionspace, \gamma, T)$.

\subsection{Representative World Models in MBRL}\label{sec:kernel_classification}

\paragraph{\textbf{Dreamer}} Dreamer~\cite{hafner2019learning, hafner2019dream, hafner2020mastering} is a general visual model-based RL method that learns world models from pixels and trains a policy via latent imagination. Specifically, Dreamer learns a latent dynamics model called Recurrent State Space Model (RSSM)~\cite{hafner2019learning}, which consists of following four components:
% \begin{gather}
% \begin{aligned}
% &\text{Representation model:} &&s_t\sim q_\theta(s_{t} \,|\,s_{t-1},a_{t-1}, o_{t}) &&\text{Image decoder:} &&\hat{o}_t\sim p_\theta(\hat{o}_{t} \,|\,s_{t}) \\
% &\text{Transition model:} &&\hat{s}_t\sim p_\theta(\hat{s}_{t} \,|\,s_{t-1}, a_{t-1}) &&\text{Reward predictor:} &&\hat{r}_t\sim p_\theta(\hat{r}_{t} \,|\,s_{t})
% \label{eq:dreamer_world_model}
% \end{aligned}
% \end{gather}

\begin{gather}
\begin{aligned}
&\text{Representation model:} &&s_t\sim q_\theta(s_{t} \,|\,s_{t-1},a_{t-1}, o_{t}) \\ &\text{Image decoder:} &&\hat{o}_t\sim p_\theta(\hat{o}_{t} \,|\,s_{t}) \\
&\text{Transition model:} &&\hat{s}_t\sim p_\theta(\hat{s}_{t} \,|\,s_{t-1}, a_{t-1}) \\ &\text{Reward predictor:} &&\hat{r}_t\sim p_\theta(\hat{r}_{t} \,|\,s_{t})
\label{eq:dreamer_world_model}
\end{aligned}
\end{gather}
The representation model extracts model state $s_{t}$ from previous model state $s_{t-1}$, previous action $a_{t-1}$, and current observation $o_{t}$. The transition model predicts the future state $\hat{s}_{t}$ using input $s_{t-1}$ and $a_{t-1}$, without access to current observation $o_{t}$. Then the image decoder reconstructs raw pixels $hat{o}_{t}$ to provide a learning signal, and the reward predictor $\hat{r}_t$ enables us to compute rewards from future model states without decoding future frames. All model parameters $\theta$ are trained to jointly learn visual representations and environment dynamics by minimizing  the negative variational lower bound (ELBO)~\cite{kingma2013auto}:
\begin{gather}
\begin{aligned}
    &\mathcal{L}(\theta) \doteq  \mathbb{E}_{q_{\theta}\left(s_{1:T}|a_{1:T},o_{1:T}\right)}\Big[ \\
    &\textstyle\sum_{t=1}^{T} \Big(
    -\ln p_{\theta}(o_{t}|s_{t})
    -\ln p_{\theta}(r_{t}|s_{t}) + \\&\beta\,\text{KL}\left[ q_{\theta}(s_{t}|s_{t-1},a_{t-1},o_{t}) \,\Vert\,  p_{\theta}(\hat{s}_{t}|s_{t-1},a_{t-1}) \right]
    \Big)\Big],
    \label{eq:dreamer_objective}
\end{aligned}
\end{gather}
where $\beta$ is a hyperparameter that controls the tradeoff between the quality of visual representation learning and the dynamics learning. Then, Dreamer uses the world model through the paradigm of planning. The critic is learned to estimate the values from pure imaginary rollouts and the actor is trained to maximize the values by propagating analytic gradients back through the transition model:
\begin{gather}
\begin{aligned}
&\text{Actor:} &&\hat{a}_{t} \sim p_{\psi}(\hat{a}_{t}\,|\,\hat{s}_{t}) \\&\text{Critic:} &&v_{\xi}(\hat{s}_{t}) \approx \mathbb{E}_{p_{\theta}}\left[\textstyle\sum_{i \leq t}\gamma^{i - t} \hat{r}_{i}\right],
\label{eq:actor_critic}
\end{aligned}
\end{gather}
where \{$\hat{s}_{t}$, $\hat{a}_{t}$, $\hat{r}_{t}$\} is imagined sequence and extracted by the dynamics model and the reward model in~\cref{eq:dreamer_world_model}. The entire reasoning and computational process is carried out in the latent space to ensure accurate predictions over time, so the latent policy input $\hat{s}_{t}$ encoded by the representation model from original raw pixel $o_t$.
Then the critic is trained to estimate the $\lambda$-target~\cite{sutton1999reinforcement} as follows:
\begin{align}
    &\mathcal{L}^{\texttt{critic}}(\xi)\doteq\mathbb{E}_{p_{\theta}}\left[\sum^{H-1}_{t=1} \frac{1}{2} \left(v_{\xi}(\hat{s}_{t}) - \text{sg}(V_{t}^{\lambda})\right)^{2}\right],
    \label{eq:critic_loss}\\
    &V_{t}^{\lambda}\doteq \hat{r}_{t} + \gamma
    \begin{cases}
      (1 - \lambda)v_{\xi}(\hat{s}_{t+1})+\lambda V_{t+1}^{\lambda} & \text{if}\ t<H \\
      v_{\xi}(\hat{s}_{H}) & \text{if}\ t=H,
    \end{cases}
    \label{eq:lambda_return}
\end{align}
where $\text{sg}$ is a stop gradient function. The actor is trained by maximizing the imagined return by back propagating the gradients through the learned transition models as follows:
\begin{align}
    \mathcal{L}^{\texttt{actor}}(\psi)\doteq \mathbb{E}_{p_{\theta}} \left[-V_{t}^{\lambda} - \eta\,\text{H}\left[a_{t}|\hat{s}_{t}\right] \right],
    \label{eq:actor_loss}
\end{align}
where the entropy of actor $\,\text{H}\left[a_{t}|\hat{s}_{t}\right]$ is used for encouraging exploration, while $\eta$ is a hyperparameter to adjusts the strength of entropy regularization.

\paragraph{\textbf{TDMPC}} TDMPC~\cite{hansen2022temporal} is an mainstream MBRL algorithm that uses world model for Model Predictive Control (MPC) and terminal value function for Temporal Difference (TD) learning. TDMPC learns world models that compresses the history of observations $o_t$ into a compact feature space variable $s_t$ and enables policy learning and planning in this space. It build a task-oriented latent dynamic Model including three major model components:
\begin{equation}
\vspace{-5pt}
\begin{array}{ll}
\text { Representation: } & \mathbf{s}_t=E_\theta\left(o_t\right) \\
\text { Latent dynamics: } & \mathbf{s}_{t+1}=D_\theta\left(\mathbf{s}_t, a_t\right) \\
\text { Reward predictor: } & \hat{r}_t \sim R_\theta\left(\mathbf{s}_t, a_t\right) 
\end{array}
\end{equation}

The representation model that encodes state $\mathbf{o_t}$ to a model state $\mathbf{s_t}$ and characterizes spatial constraints by latent consistency loss, then the latent dynamics model that predicts future model states $\mathbf{s}_{t+1}$ without reconstruction. Then, TDMPC builds actor and critic based on DDPG~\cite{lillicrap2015continuous}:
\begin{equation}
\begin{array}{ll}
\text { Critic: }  \hat{q}_t=Q_\theta\left(s_t, a_t\right) \qquad
\text { Actor: } \hat{a}_t \sim \pi_\phi\left(s_t\right)
\end{array},
\end{equation}
All model parameters $\theta$ except actor are jointly optimized by minimizing the temporally objective:
% \begin{equation}
% \label{eq:model loss}
% \footnotesize
% \begin{aligned}
% &\mathcal{L}\left(\theta, \mathcal{D}\right)=\mathbb{E}_{\left(o, \mathbf{a}, s_t, r_t\right)_{t:t+H} \sim \mathcal{D}} \left[
% \sum_{i=t}^{t+H}\left( c_1 \underbrace{\left\|R_\theta\left(s_i, \mathbf{a}_i\right)-r_i\right\|_2^2}_{\text {Reward prediction}}\\
% &+c_2 \underbrace{\left\|D_\theta\left(s_i, \mathbf{a}_i\right)-E_{\theta-}\left(o_{i+1}\right)\right\|_2^2}_{\text{Latent state consistency}} \right. \right.\\
% &\left.\left.+c_3 \underbrace{\left\|Q_\theta\left(s_i, \mathbf{a}_i\right)-\left(r_i+\gamma Q_{\theta^{-}}\left(s_{i+1}, \pi_\theta\left(s_{i+1}\right)\right)\right)\right\|_2^2}_{\text {Value prediction  }}\right)\right].
% \end{aligned}
% \end{equation}

\begin{equation}
\label{eq:model_loss}
\small
\begin{aligned}
\mathcal{L}(\theta, \mathcal{D}) = &\mathbb{E}_{(o, \mathbf{a}, s_t, r_t)_{t:t+H} \sim \mathcal{D}} \Bigg[  \sum_{i=t}^{t+H} \Big( c_1 \left\| R_\theta(s_i, \mathbf{a}_i) - r_i \right\|_2^2 \\
& + c_2 \left\| D_\theta(s_i, \mathbf{a}_i) - E_{\theta-}(o_{i+1}) \right\|_2^2 \\
& + c_3 \left\| Q_\theta(s_i, \mathbf{a}_i) - \left( r_i + \gamma Q_{\theta^{-}}(s_{i+1}, \pi_\theta(s_{i+1})) \right) \right\|_2^2 \Big) \Bigg].
\end{aligned}
\end{equation}
Hyper-parameters $c_i$ are adjusted to balance the loss function and a trajectory of length $H$ is sampled from the replay buffer $\mathcal{D}$.

\paragraph{\textbf{MBPO}} MBPO~\cite{DBLP:conf/nips/JannerFZL19} is dyna-style MBRL algorithm~\cite{sutton1999reinforcement} which similarly maintains dynamics $s_{t+1}=D_\theta\left(s_t, a_t\right)$ and reward models $r_{t+1}=R_\theta\left(s_t, a_t\right)$. The policy $\pi_\phi$ is optimized using real environmental data and more model-generated data. 

In general, although there are different ways to build and use models, the core of the world model remains a dynamics model, i.e., predicting state transitions in the environment, which can be roughly divided into two categories:
\textbf{with non-recurrent kernel} and \textbf{with recurrent kernel}. The formal definition of both kernels are as follows:
$$ h_t=\left\{
\begin{array}{rcl}
f(s_{t-1}, a_{t-1})         &      & \text{With non-recurrent kernel}\\
f(h_{t-1}, s_{t-1}, a_{t-1})&      & \text{With recurrent kernel}
\end{array} \right. $$
Non-recurrent kernel~(MBPO, TDMPC) are relatively basic kernel for modeling, which are often implemented as Fully-Connected Networks. Non-recurrent kernel takes the current state $s_{t-1}$ and action $a_{t-1}$ as input, outputs the latent state prediction $h_t$. Compare to non-recurrent kernel, recurrent kernel~(Dreamer) is implemented as RNN and takes the additional input $h_{t-1}$, which performs better under the POMDP setting. For both kernels, the $s_t$ and $r_t$ can be generated from the latent prediction $h_t$.

\section{Environment Dynamics Decomposition}

\subsection{Motivation}\label{sec:motivation}

An accurate world model is critical in MBRL policy deriving. To decrease the model error, existing works propose various techniques as introduced in \cref{section:related_work}. However, these techniques improve environment modeling in a black-box manner, which ignores the inner properties of environment dynamics and kinematics, resulting in inaccurate world model construction and poor policy performance. To address this problem, we propose two important environment properties when modeling an environment:

\begin{itemize}
\item \textbf{Decomposability}:
  The environment dynamics can be decomposed into multiple sub-dynamics in various ways and the decomposed sub-dynamics can be combined to reconstruct the entire dynamics.
\item \textbf{Traceability}:
  The environment dynamics can be traced to the action's impact, and each sub-dynamics can be traced to the impact caused by a part of the action.
\end{itemize}

As shown in \cref{fig:motivation_example}, we demonstrate the decomposability: we can decompose the dynamics into $\{\textit{hip, }, \textit{knee}, \textit{ankle}\}$ sub-dynamics or $\{\textit{left}, \textit{right}\}$ sub-dynamics, which depends on the different decomposition perspectives and the combination of decomposed sub-dynamics can constitute the entire dynamics. Then we explain the traceability: each sub-dynamics can be traced to the corresponding subset of action dimensions:
for the $\textit{hip}$ dynamics, it can be regarded as the impact caused by the $\textit{left-thigh}$ and $\textit{right-thigh}$ action dimensions.
The above two properties are closely related to environment modeling: the decomposability reveals the existence of sub-dynamics, which allows us to model the dynamics separately, while the traceability investigates the causes of the dynamics and guides us to decompose the dynamics at its root (i.e. the action). 

Intuitively, it is easy to think of a decomposition of the state space~\cite{DBLP:journals/neco/DoyaSKK02}. For example, the state from the Walker2D domain consists of positional values and velocities of different body parts of the Walker, so we can simply decompose the state space. However, they are limited to low-dimensional state input, and it is difficult to decompose uncontrollable elements such as environmental information. We adopt a novel perspective to implicitly consider robot dynamics and kinematics, analyzing the cause of dynamics and decomposing it from its root: the action space, which makes the modeling of environmental dynamics more reasonable and
scalable. We discuss the different perspectives on decomposition in more depth in \cref{sec:D2P_comparison}.

\begin{figure*}[t]
\centering
\includegraphics[width=0.98\linewidth]{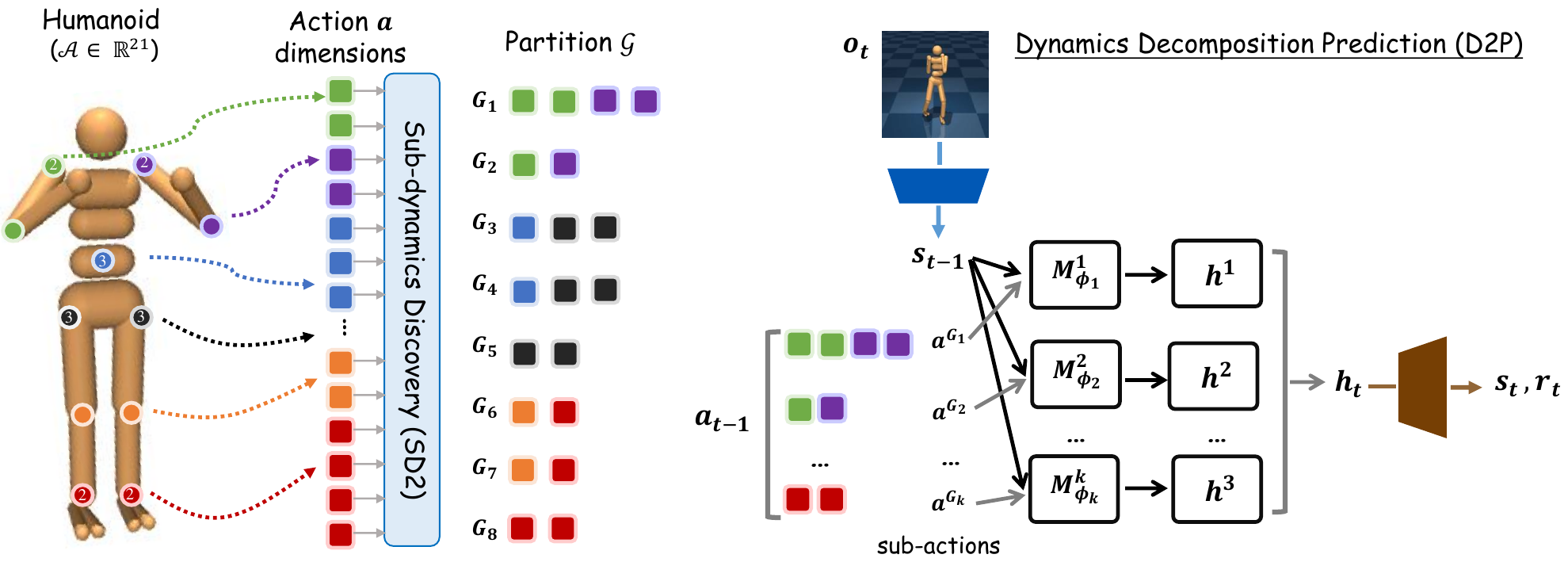}
\caption{Overview of the \alg 
 Framework. \alg contains two components: SD2 and D2P. SD2 decomposes the dynamics by generating a partition $\mathcal{G}$ on action dimensions. D2P decomposes the action $a_{t-1}$ into multiple sub-actions according to $\mathcal{G}$ and makes decomposing predictions based on $s_t$ and each sub-action $a^{G_k}$. The prediction $h_t$ is the combined output of all sub-dynamics models, from which the next state $s_{t}$ and reward $r_{t}$ are generated.}
\label{fig:ED2_overview}
\end{figure*}

To take the above properties into account, we propose the environment dynamics decomposition (\alg) framework (as shown in \cref{fig:ED2_overview}), which contains two key components: sub-dynamics discovery (SD2) and dynamics decomposition prediction (D2P).
More specifically, by considering the traceability, we propose to discover the latent sub-dynamics by analyzing the action (SD2, \cref{fig:ED2_overview}~(left); by considering the decomposability, we propose building the world model in a decomposing manner (D2P, \cref{fig:ED2_overview}~(right)).
Our framework can be used as a backbone in MBRL and the combination can lead to performance improvements over existing MBRL algorithms.

\subsection{Dynamics Decomposition Prediction}\label{sec:D2P}
Given an environment with a $m$-dimensional action space $A \subset \mathbb{R}^m$, the index of each action dimension constitutes a set $\Lambda = \{1, 2, \cdots, m \}$, any disjoint partition $\mathcal{G} = \{G_{1}, \dots, G_{k}\}$ over $\Lambda$ corresponds to a particular way of decomposing the action space. For each action dimension $i$ in $\Lambda$, we define the action space as $A^i$, which satisfies $A = A^1 \times \cdots \times A^m$. The decomposition of the action space under partition $\mathcal{G}$ is defined as $A^\mathcal{G} = \{A^{G_1}, \cdots, A^{G_k}\}$, where sub-action space $A^{G_j} = \prod_{x \in G_j} A^x$. Based on the above definitions,
we define the dynamics decomposition for $P$ under partition $\mathcal{G}$ as follows:
% the environment dynamics function $P$ can be approximated in a decomposed fashion with sub-dynamics as follow:
\begin{defi}
Given a partition $\mathcal{G}$,
the decomposition for $P : S \times A \rightarrow S$ can be defined as:
\begin{equation}
\label{equation:dyna_decomp}
    \begin{aligned}
            P(s,a) = f_c \left(\frac{1}{k} \sum_{i=1}^k P_i(s, a^{G_i}) \right), \forall s,a \in S \times A,
    \end{aligned}
\end{equation}
with a set of sub-dynamics functions $\{P_1, ..., P_k\}$ that $P_i : S \times A^{G_i} \rightarrow H$, and a decoding function $f_c: H \rightarrow S$.
Note $H$ is a latent space and $a^{G_i} \in A^{G_i}$ is a sub-action (projection) of action $a$.
\end{defi}

Intuitively, the choice of partition $\mathcal{G}$ is significant to the rationality of dynamics decomposition, which should be reasonably derived from the environments.
In this section, we mainly focus on dynamics modeling, and we will introduce how to derive the partition $\mathcal{G}$ by using SD2 in \cref{sec:dynamics-discovery}.

To implement D2P, we use model $M^i_{\phi_i}$ parameterized by $\phi_i$ to approximate each sub-dynamics $P_i$.
As illustrated in \cref{fig:ED2_overview},
given a partition $\mathcal{G}$, an action $a$ is divided into multiple sub-actions $\{a^{G_1}_t, \cdots, a^{G_k}_t\}$,
each model $M^i_{\phi_i}$ takes state $s_{t-1}$ and the sub-action $a^{G_i}_{t-1}$ as input and output a latent prediction $h^{i} \in H$.
The separate latent predictions $\{h^1, \cdots, h^k\}$ are aggregated and then decoded for the generation of state $s_{t}$ and reward $r_{t}$.
For each kernel described in \cref{sec:kernel_classification}, we provide the formal description here when combine with D2P:

\small
$$ h_t=\left\{
\begin{array}{rcl}
\frac{1}{k} \sum_{i=1}^k f(s_{t-1}, a^{G_i}_{t-1})&& \text{For non-recurrent kernel}\\
\frac{1}{k} \sum_{i=1}^{k} f(h_{t-1}, s_{t-1}, a^{G_i}_{t-1})&&\text{For recurrent kernel}
\end{array} \right. $$

\begin{figure}
\centering
% \vspace{-0.8cm}
\includegraphics[width=0.99\linewidth]{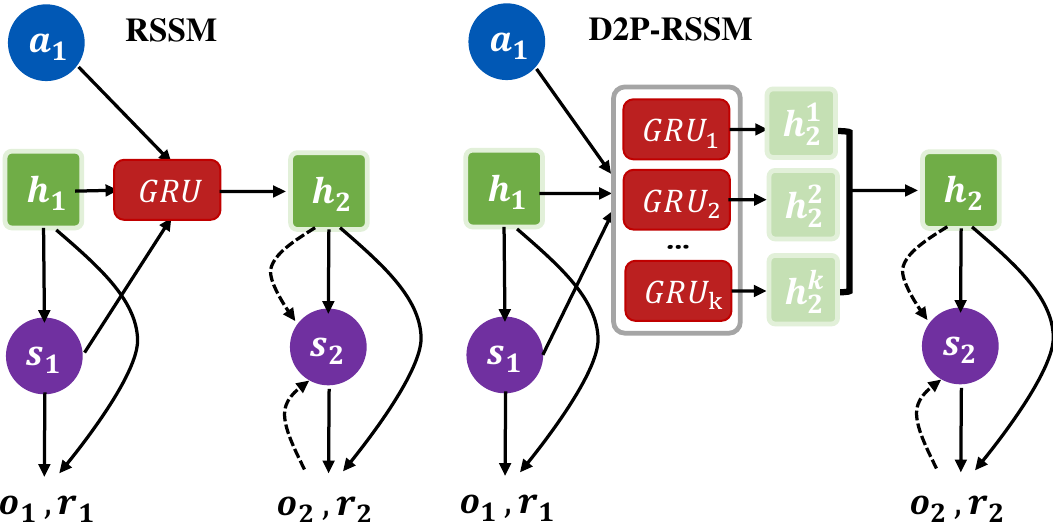}
\caption{Extension of RSSM with D2P.}
\label{fig:RSSM-ED2}
\end{figure}

We propose a set of kernels, where each kernel builds a specific sub-dynamics with the input of current state $s$, corresponding sub-action $a^{G_i}$ and hidden state $h_{t-1}$ (ignored when applying on non-recurrent kernel). The output of all kernels is averaged to get the final output $h_t$. The prediction of reward $r_t$ and state $s_t$ is generated from output $h_t$. Specifically, we provide an example when combining with the kernel of Recurrent State-Space Model (RSSM) \cite{DBLP:conf/icml/HafnerLFVHLD19} in \cref{fig:RSSM-ED2}, which is a representative recurrent kernel-based world model. The original single kernel implemented as GRU are replace by multiple kernels with different action input.

% \textbf{Decomposed non-recurrent kernel:} With considering multiple sub-dynamics, we propose to model each sub-dynamics with a distinct kernel. The formal description is as follows:
% \begin{equation}
%     h_t = \frac{1}{k} \sum_{i=1}^k f(s_{t-1}, a^{G_i}_{t-1})\label{equ:non_rec}
% \end{equation}
% We propose a set of kernels, where each kernel models a specific sub-dynamics with the input of current state $s$ and corresponding sub-action $a^{G_i}$. the output of all kernels is averaged to get the final output $h_t$.
% The prediction of reward $r_t$ and state $s_t$ is generated from the output $h_t$.

% \textbf{Decomposed recurrent kernel:} To apply D2P on POMDP, we also combined recurrent kernel with D2P. The formal description is as follows:
% \begin{equation}
%     h_t = \frac{1}{k} \sum_{i=1}^{k} f(h_{t-1}, s_{t-1}, a^{G_i}_{t-1})\label{equ:rec}
% \end{equation}
% Recurrent State-Space Model (RSSM) \cite{DBLP:conf/emnlp/ChoMGBBSB14} is a representative recurrent kernel-based world model and we take it as an example. As shown in Figure~\cref{fig:RSSM-ED2}, we combine RSSM with D2P by replacing the single GRU network with multiple GRU networks. With the current state $s_{t-1}$, current hidden state $h_{t-1}$ and sub-action $a^{G_i}_{t-1}$, GRU network $GRU_i$ outputs the hidden state $h^i_t$. All generated hidden states are aggregated to the final hidden state $h_t$, which can be either used for $s_t$ generation and $h_{t+1}$ prediction.

\subsection{Sub-dynamics Discovery}\label{sec:dynamics-discovery}
The traceability of the environment introduced in \cref{sec:motivation} provides us with a basis for dynamics decomposition: the decomposition on dynamics can be converted to the decomposition on the action space. Therefore, we present the SD2 module for the decomposition of action space and discuss three implementations:

\paragraph{Complete decomposition~(CD)} Complete decomposition regards each action dimensions as a sub-dynamics. With the Walker task in \cref{sec:motivation} as the example: the straightforward SD2 implementation is the complete decomposition, which regards each action dimension as a sub-dynamics and decomposes the dynamics completely. Specifically, complete decomposition decomposes the dynamics into six sub-dynamics: $\{Left, Right\} \times \{Hip, Knee, Ankle\}$. However, complete decomposition ignores the inner action dimensions correlations, which limits its performance in many tasks. For example, the three action dimensions $\{Left\_Hip, Left\_Knee, Left\_Ankle\}$ affect the dynamics of the front part together, thus simply separate these action dimensions would affect the prediction accuracy. Besides, complete decomposition is not scalable when the action dimension is large. 

\paragraph{Human prior~(HP)} To include the action dimension correlations, incorporating human prior for the action space decomposition is an improved implementation. Based on different human prior, we can decompose the dynamics in different ways as introduced in \cref{fig:motivation_example}. Nevertheless, although human prior considers the action dimension correlations, it is highly subjective and might lead to sub-optimal results due to the limited understanding of tasks (we also provide the corresponding experiment in \cref{sec:RQ4}). Therefore, human prior is not applicable in complex systems which is beyond human understanding.

\paragraph{Clustering~(CL)} To better discover the sub-dynamics and eliminate the dependence on human prior, 
we propose to automatically decompose the action space using the clustering-based method.
% we propose to use a clustering-based method (CL).
% To automatically discover the correlations between action dimensions, and further discover a better action space decomposition, we propose to use a clustering method.
%, which overcome the shortcoming of above two methods.
The clustering-based method contains two components: \textbf{feature extraction} and \textbf{clustering criterion}. 
Feature extraction extracts the properties of action dimension $a^i$ into feature vector $F^i$. Then we regard each action dimension as a cluster and aggregate related action dimensions together with the clustering criterion.
The effectiveness of the clustering-based method depends on the quality of feature extraction and the validity of clustering criteria, which may be different in different environments. Therefore, although we provide a general implementation later, we still suggest readers design suitable clustering-based methods according to task-specific information.

\textbf{Feature Extraction:} We extract the properties of each action dimension by computing the Pearson correlation coefficient between action dimensions and state dimensions. Specifically, we define the feature vector as $F^i=\langle |f^{i,1}|,\cdots,|f^{i,n}| \rangle$, where each $f^{i,j}$ denotes the Pearson correlation coefficient between action dimension $i$ and state dimension $j$. $F^i$ describes the impact caused by action dimension $i$ and
$f^{i,j}$ is calculated by the corresponding action value $a^i$ and state value changes $\Delta s^j$ (which is the difference between the next state and the current state):
\begin{equation} \label{alg:f}
    f^{i,j} = \frac{cov(a^i, \Delta s^j)}{\sigma_{a^i}\sigma_{\Delta s^j}}
\end{equation}
where $cov$ denotes the covariance and $\sigma$ denotes the standard deviation. 

\textbf{Clustering Criterion:}
We define the clustering criterion as the relationship between clusters, which can be formalized as follow:

% \begin{align} \label{alg:cri}
%     Rela(G_i, G_j) = & R(G_i, G_j) -
%     & \frac{R(G_j, G_{-i}) \times \omega^{j, -i} + R(G_i, G_{-j}) \times \omega^{i, -j}}{\omega^{i, -j} + \omega^{j, -i}}
% \end{align}

\begin{equation}
\label{alg:cri}
\begin{split}
\text{Rela}(G_i, G_j) = & R(G_i, G_j) - \\
    & \frac{R(G_j, G_{-i}) \times \omega^{j, -i} + R(G_i, G_{-j}) \times \omega^{i, -j}}{\omega^{i, -j} + \omega^{j, -i}}
\end{split}
\end{equation}

where $G_{-i} = \Lambda \setminus G_i$, $\omega^{i, j}=|G_i|\times|G_j|$ and $R(G_i, G_j) = - \frac{1}{\omega^{i, j}} \sum_{A^i \in G_i} \sum_{A^j \in G_j} ||F^i, F^j||_D$.
$||\cdot||_D$ measures the distance between vectors under distance function $D$ (we choose the negative cosine similarity as $D$).

\begin{algorithm}[H]
\small
\caption{Selectable clustering-based method.}
\label{alg:algorithm}
\textbf{Input}: Task $\mathcal{E}$, clustering threshold $\eta$
\begin{algorithmic}[0]
\STATE Initialize cluster set $\mathcal{G}=\{\{1\},\cdots,\{m\}\}$ according to $\mathcal{E}$, a random policy $\pi_{rand}$, dataset $\mathcal{D}_c \to \varnothing$
\FOR{$i = 1,2,\cdots,T$}
\STATE Collect and store samples in $\mathcal{D}_c$ with $\pi_{rand}$ \\
\ENDFOR
\STATE Calculate $F^i$ for each action dimension $i$ with $\mathcal{D}_c$\\
\WHILE{$|\mathcal{G}|>1$}
% \STATE \yan{//select clustering elements}\\
\STATE $G_{max_1}, G_{max_2} = \argmax_{G_i, G_j \in \mathcal{G}} Rela(G_i, G_j)$\\
\IF{$Rela(G_{max_1}, G_{max_2}) > \eta$}
    \STATE Remove $G_{max_1}$ and $G_{max_2}$ from $\mathcal{G}$\\
    \STATE Add $G_{max_1} \cup G_{max_2}$ to $\mathcal{G}$\\
\ELSE
    \STATE Stop clustering\\
\ENDIF
\ENDWHILE
\RETURN $\mathcal{G}$\\
\end{algorithmic}
\end{algorithm}

\cref{alg:algorithm} presents the overall implementation of the clustering-based method.
As \cref{alg:algorithm} describes, with input task $\mathcal{E}$ and clustering threshold $\eta$, we first initialize the cluster set $\mathcal{G}$ containing $m$ clusters (each for a single action dimension), a random policy $\pi_{rand}$, and an empty dataset $\mathcal{D}_c$. Then for $T$ episodes, $\pi_{rand}$ collects samples from the environment and we calculate $F^i$ for each action dimension $i$. After that, for each clustering step, we select the two most relevant clusters from $\mathcal{G}$ and cluster them together.
The process ends when there is only one cluster, or when the correlation of the two most correlated clusters is less than the threshold $\eta$. $\eta$ is a hyperparameter which assigned with a value around 0 and empirically adjusted. 

\subsection{ED2 for MBRL Algorithms}

\begin{algorithm}[H]
\small
\caption{\alg-Dreamer}
\label{alg:ED2-Dreamer}
\textbf{Input}: Task $\mathcal{E}$, clustering threshold $\eta$
\begin{algorithmic}[0]
\STATE \emph{// Sub-dynamics Discovery (SD2) Phase:} \\
\STATE $\mathcal{G} \leftarrow$ SD2 methods $\left( \mathcal{E},\eta \right)$\\ \emph{// Dynamics Decomposition Prediction (D2P) Phase:}
\FOR{$i = 1,2,\cdots,|\mathcal{G}|$}
\STATE Build sub-dynamics model: 
$ M^i_{\phi_i} = f(h_{t-1}, s_{t-1}, a^{G_i}_{t-1})$
\ENDFOR
\STATE Combining all sub-dynamics models:
$ M_c = \frac{1}{|\mathcal{G}|} \sum_{i=1}^{|\mathcal{G}|} M^i_{\phi_i}$

\STATE Combining $M_c$ with a decoding network $f^d_{\phi_d}$ and construct the \alg-combined world model:
$p_{\phi} = f^d_{\phi_d} (M_c)$ \\ \emph{// Training Phase:}% using \alg-combined world model and MBPO}
\STATE Initialize policy $\pi_\theta$, model: $p_{\phi}$
\STATE Optimize policy with Dreamer: $\pi_{\hat{\theta}} = \text{Dreamer}(\mathcal{E}, \pi_\theta, p_{\phi})$
\end{algorithmic}
\end{algorithm}

ED2 is a general framework and can be combined with any existing MBRL algorithms. We performed the \alg with different types of MBRL algorithms, including Dreamer~\cite{DBLP:conf/iclr/HafnerLB020, hafner2020mastering}, MBPO~\cite{DBLP:conf/nips/JannerFZL19} and TDMPC~\cite{hansen2022temporal}. Here we mainly provide the practical combination implementations of \alg with Dreamer~(\cref{alg:ED2-Dreamer}) as an example, and the other baselines are simply adapted to the corresponding dynamics model. The whole process of ED2-Dreamer contains three phases: 1) SD2 decomposes the environment dynamics of task $\mathcal{E}$, which and can be implemented by three decomposing methods introduced in Section \ref{sec:dynamics-discovery}; 2) D2P models each sub-dynamics separately and constructs the \alg-combined world model $p_{\phi}$ by combining all sub-models with a decoding network $f^d_{\phi_d}$; 3) The final training phase initializes the policy $\pi_\theta$ and the world model $p_{\phi}$, then derive the policy from Dreamer with input $\pi_\theta$, $p_{\phi}$ and task $\mathcal{E}$.

\begin{figure*}[t]
\begin{minipage}[t]{\linewidth}
\centering

\begin{minipage}[t]{0.1\linewidth}
\includegraphics[width=\linewidth]{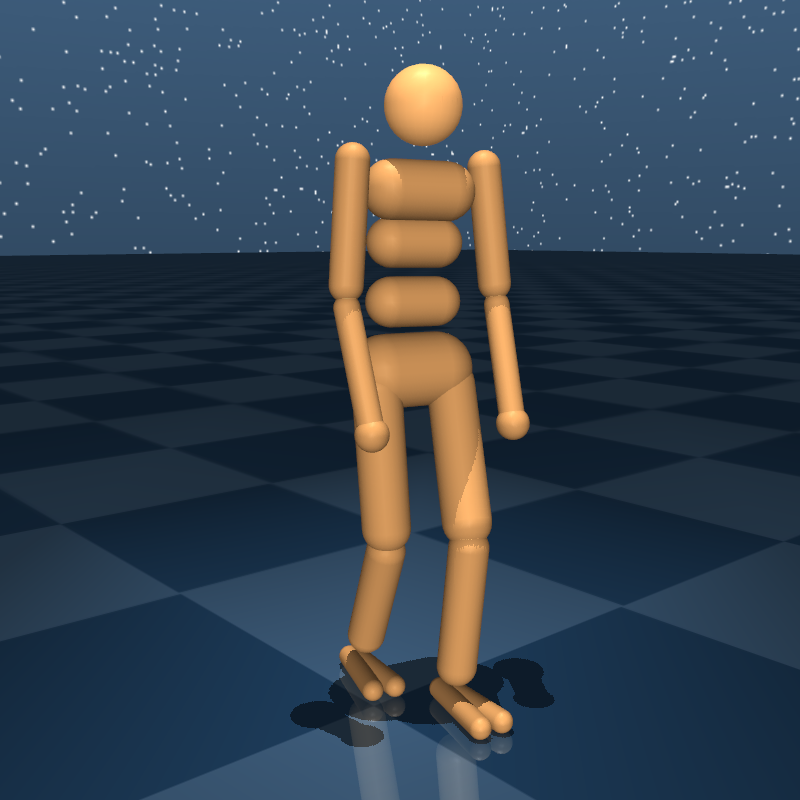}
\end{minipage}
\begin{minipage}[t]{0.1\linewidth}
\includegraphics[width=\linewidth]{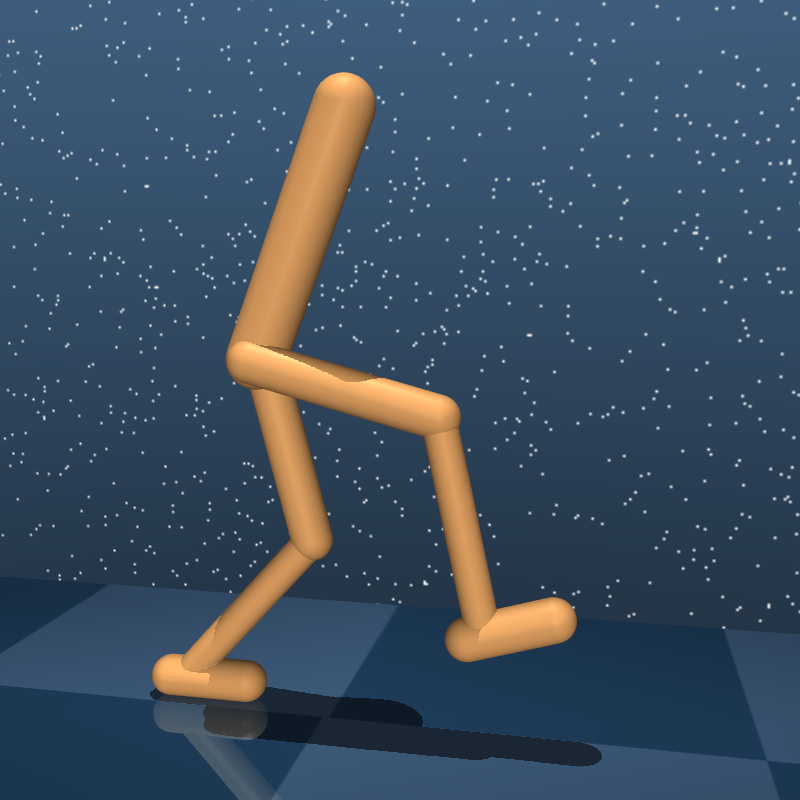}
\end{minipage}
\begin{minipage}[t]{0.1\linewidth}  \includegraphics[width=\linewidth]{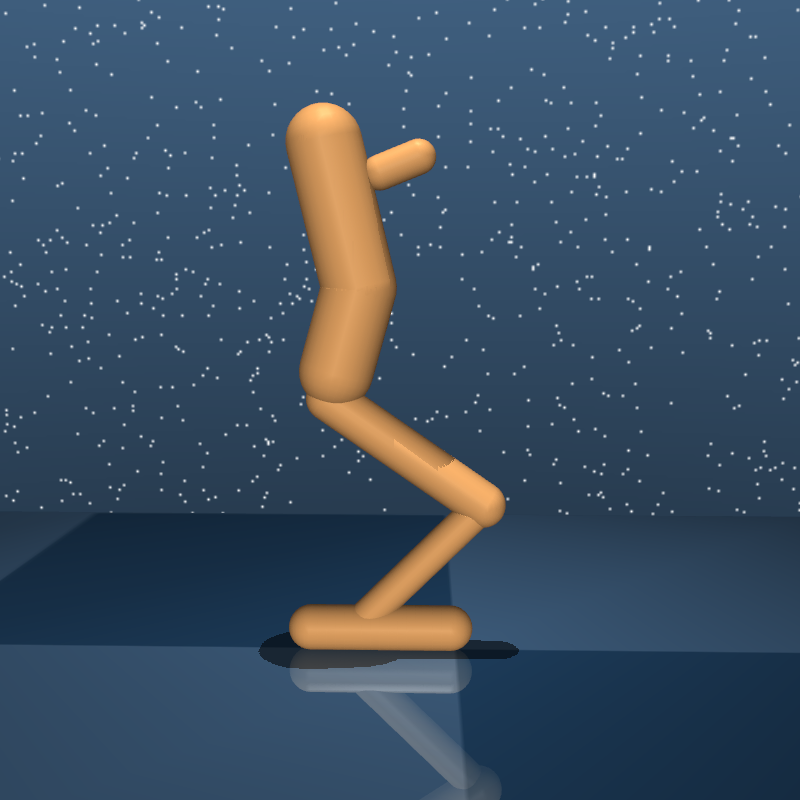}
\end{minipage}
\begin{minipage}[t]{0.1\linewidth}
\includegraphics[width=\linewidth]{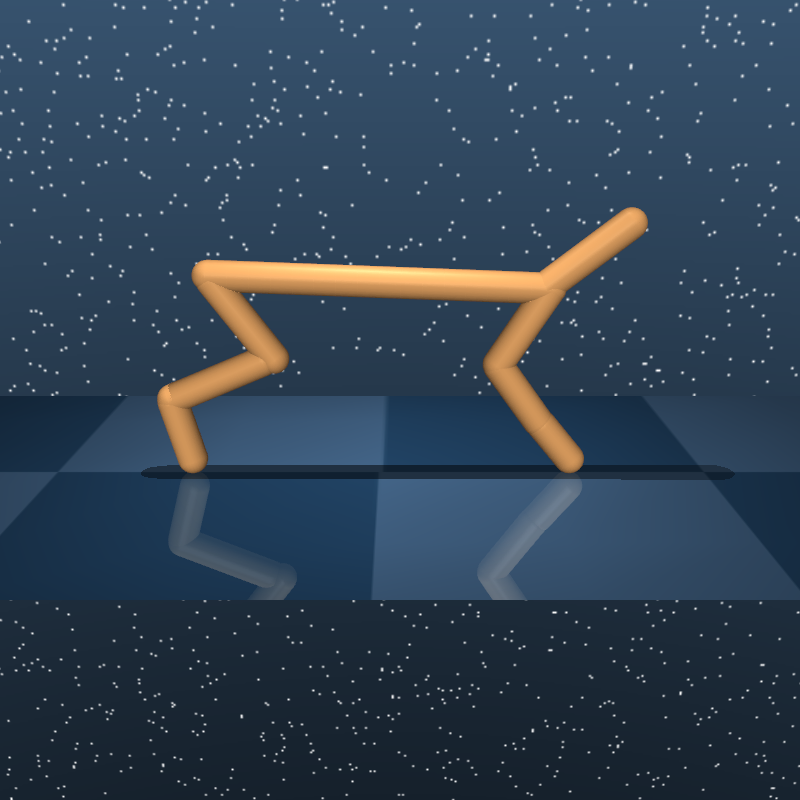}
\end{minipage}
\begin{minipage}[t]{0.1\linewidth}
\includegraphics[width=\linewidth]{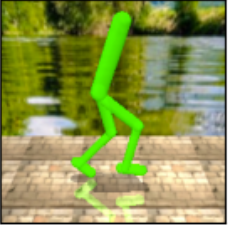}
\end{minipage}
\begin{minipage}[t]{0.1\linewidth}
\includegraphics[width=\linewidth]{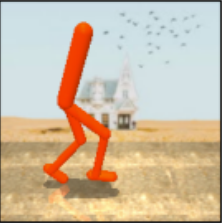}
\end{minipage}
\begin{minipage}[t]{0.1\linewidth}
\includegraphics[width=\linewidth]{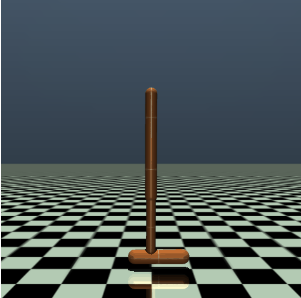}
\end{minipage}
\begin{minipage}[t]{0.1\linewidth}
\includegraphics[width=\linewidth]{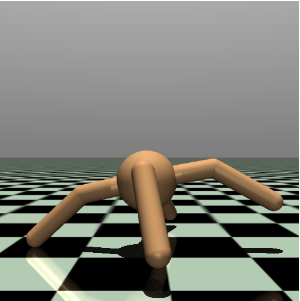}
\end{minipage}
\begin{minipage}[t]{0.1\linewidth}
\includegraphics[width=\linewidth]{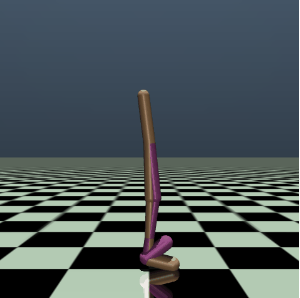}
\end{minipage}
\caption{Example image of continuous control benchmark tasks: Deepmind Control (DMC), DMControl Remastered~(DMCR), and Mujoco (left to right).}
\label{fig:benchmarks}
\end{minipage}
\end{figure*}

\begin{figure*}[t]
\begin{minipage}[t]{\linewidth}
\centering
\begin{minipage}[t]{0.31\linewidth}
\includegraphics[width=\linewidth]{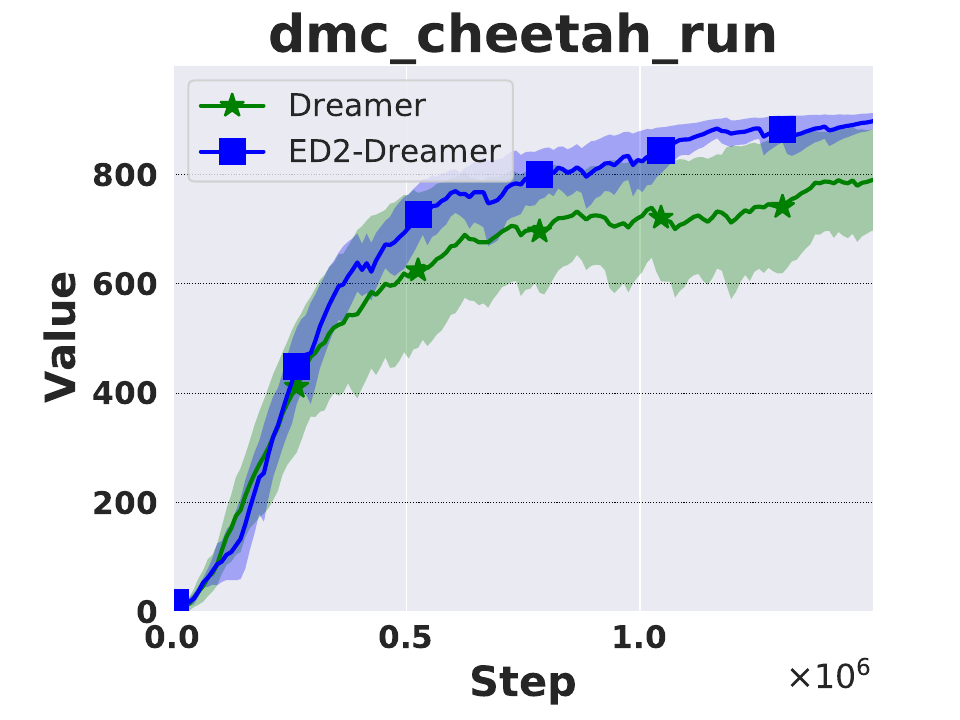}
\end{minipage}
\begin{minipage}[t]{0.31\linewidth}  \includegraphics[width=\linewidth]{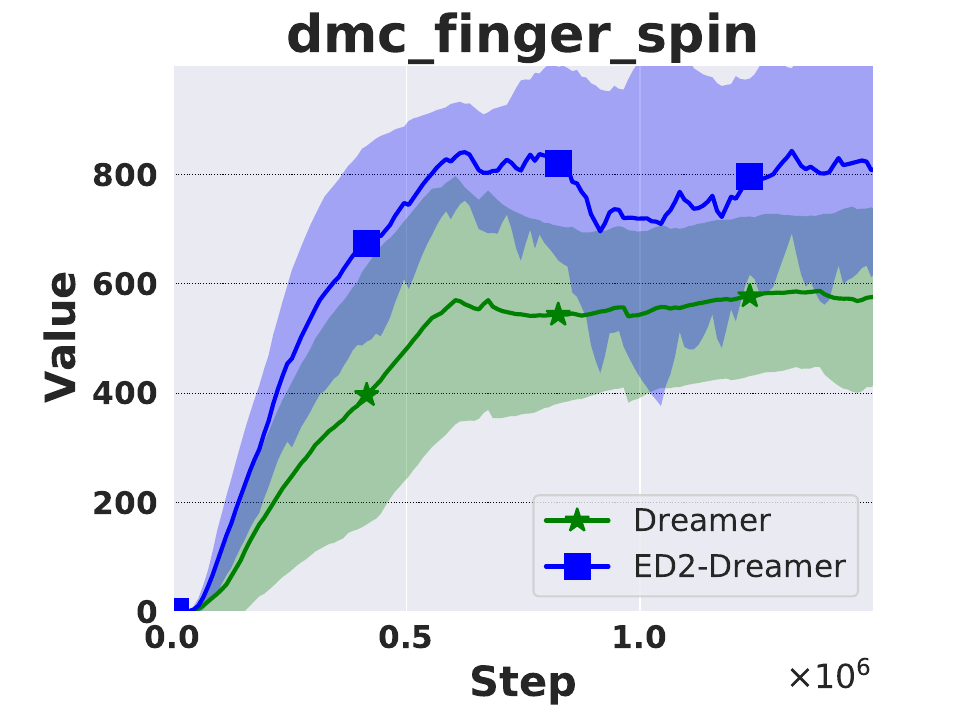}
\end{minipage}
\begin{minipage}[t]{0.31\linewidth}
\includegraphics[width=\linewidth]{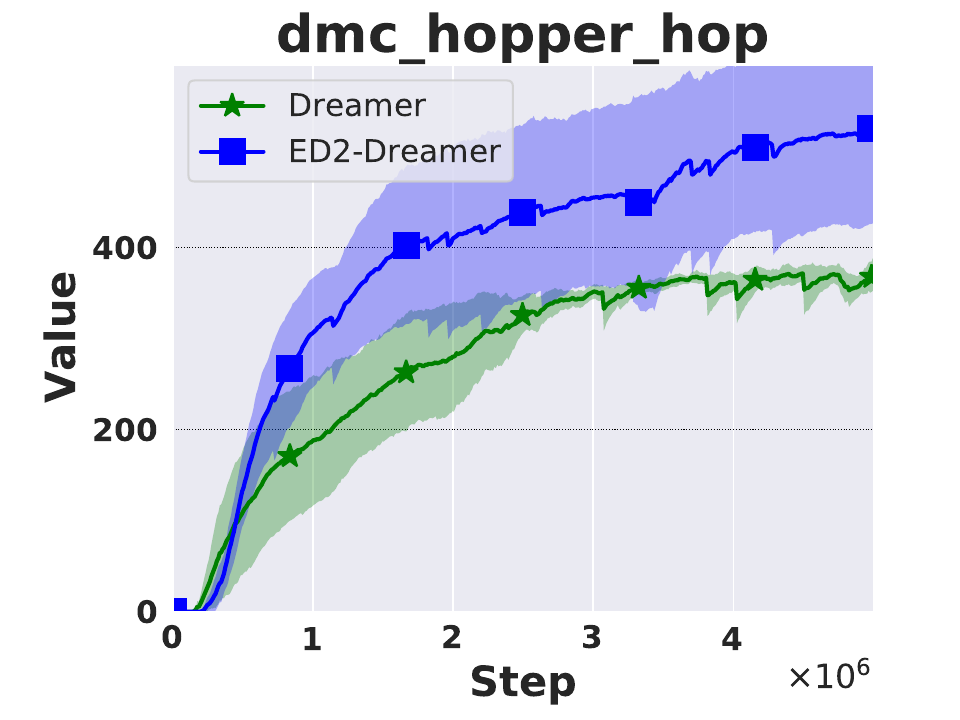}
\end{minipage}
% \begin{minipage}[t]{0.24\linewidth}
% \includegraphics[width=\linewidth]{figures/dmc_humanoid_stand_cl.pdf}
% \end{minipage}
\vfill
% \begin{minipage}[t]{0.24\linewidth}
% \includegraphics[width=\linewidth]{figures/dmc_humanoid_walk_cl.pdf}
% \end{minipage}
\begin{minipage}[t]{0.31\linewidth}
\includegraphics[width=\linewidth]{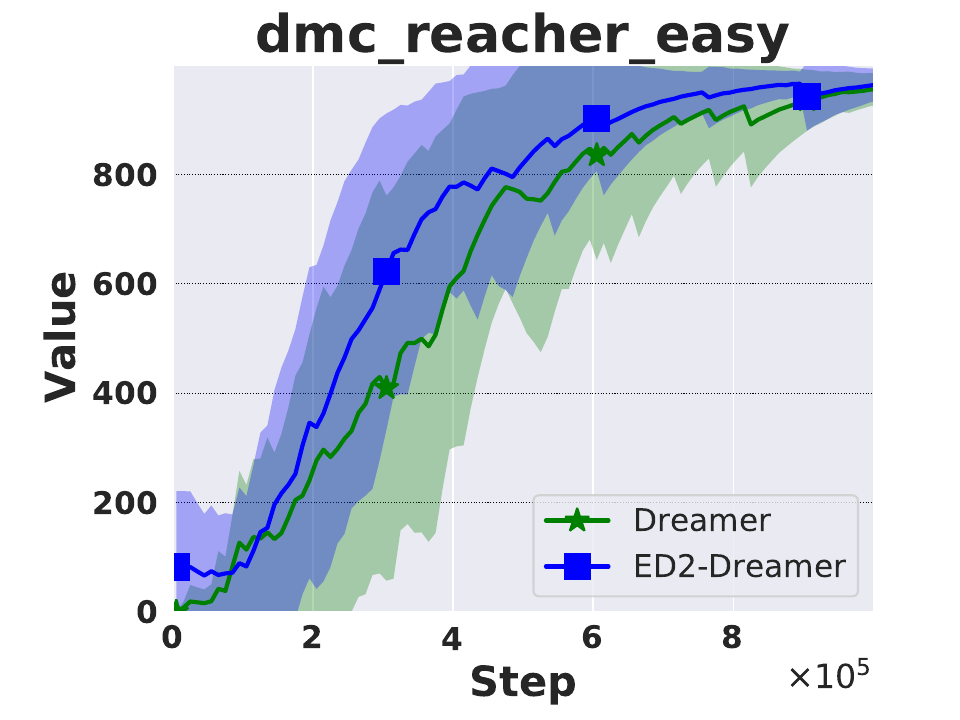}
\end{minipage}
\begin{minipage}[t]{0.31\linewidth}
\includegraphics[width=\linewidth]{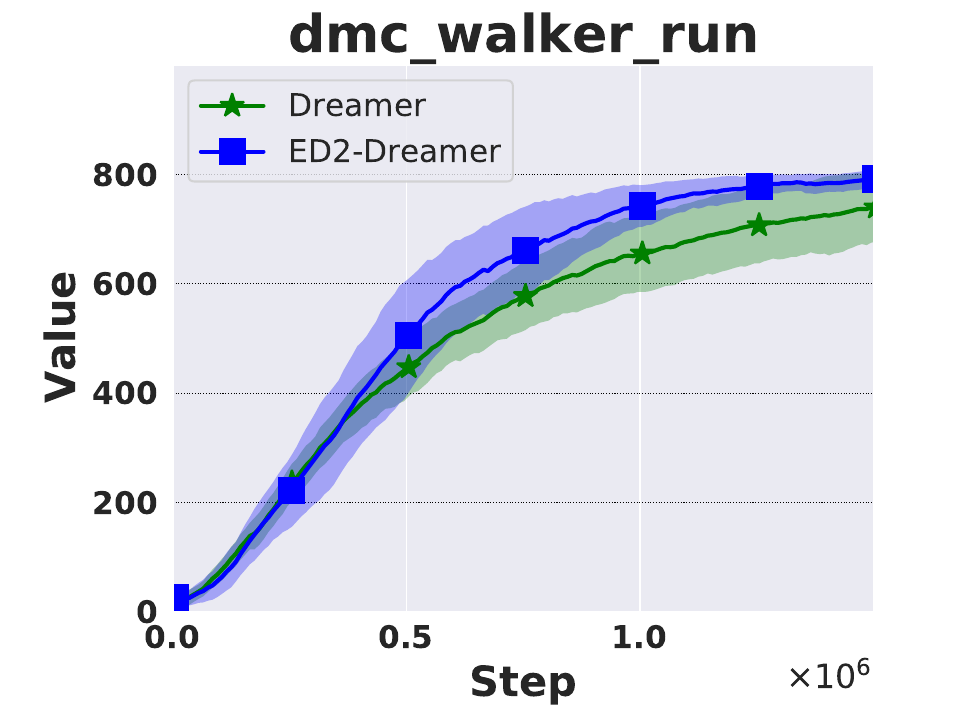}
\end{minipage}
\begin{minipage}[t]{0.31\linewidth}
\includegraphics[width=\linewidth]{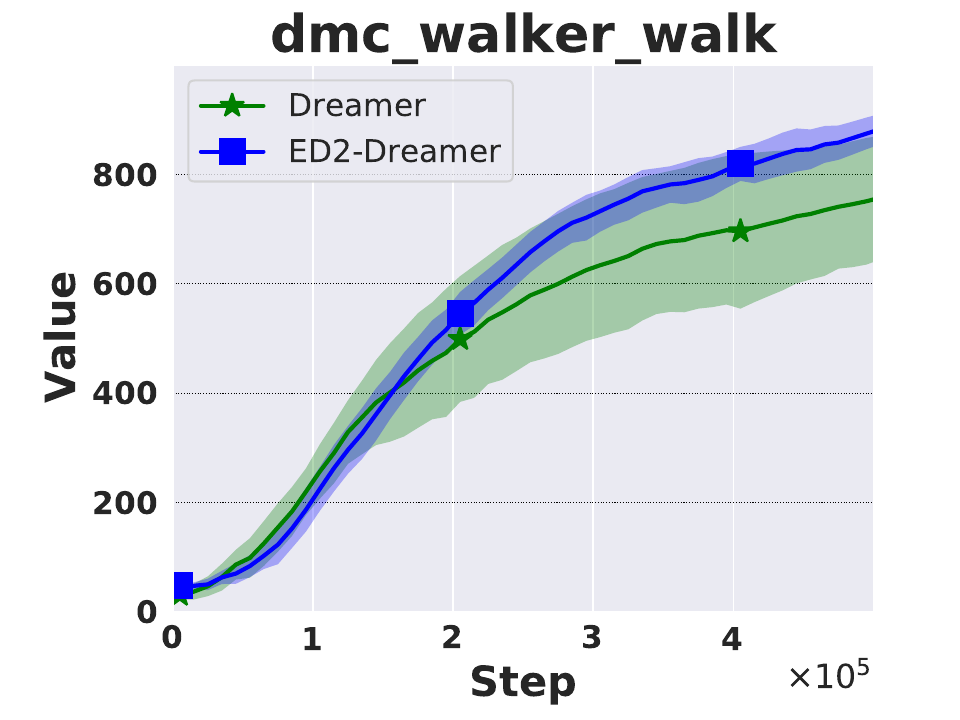}
\end{minipage}
\vfill

\begin{minipage}[t]{0.3\linewidth}
\includegraphics[width=\linewidth]{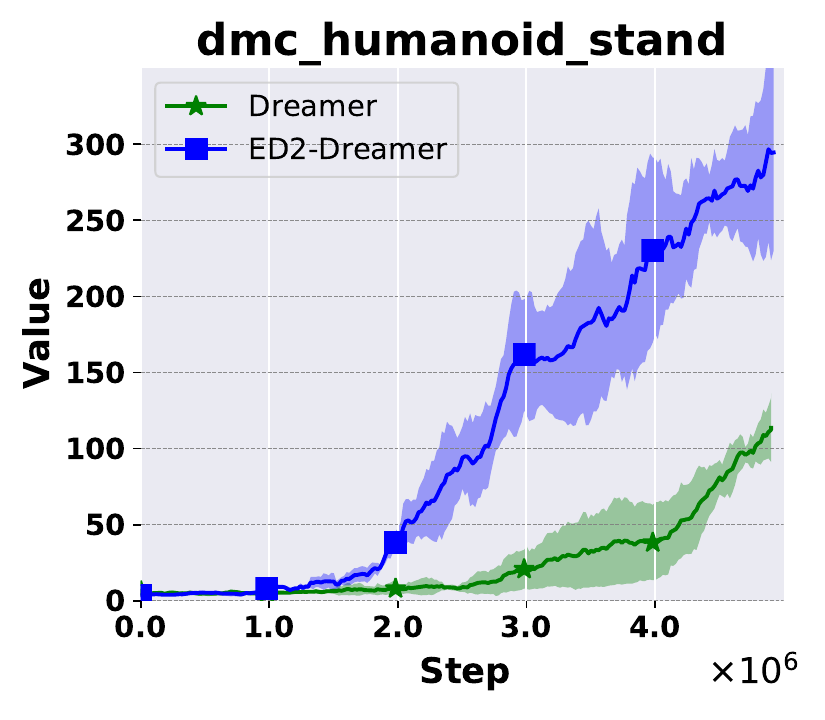}
\end{minipage}
\begin{minipage}[t]{0.3\linewidth}
\includegraphics[width=\linewidth]{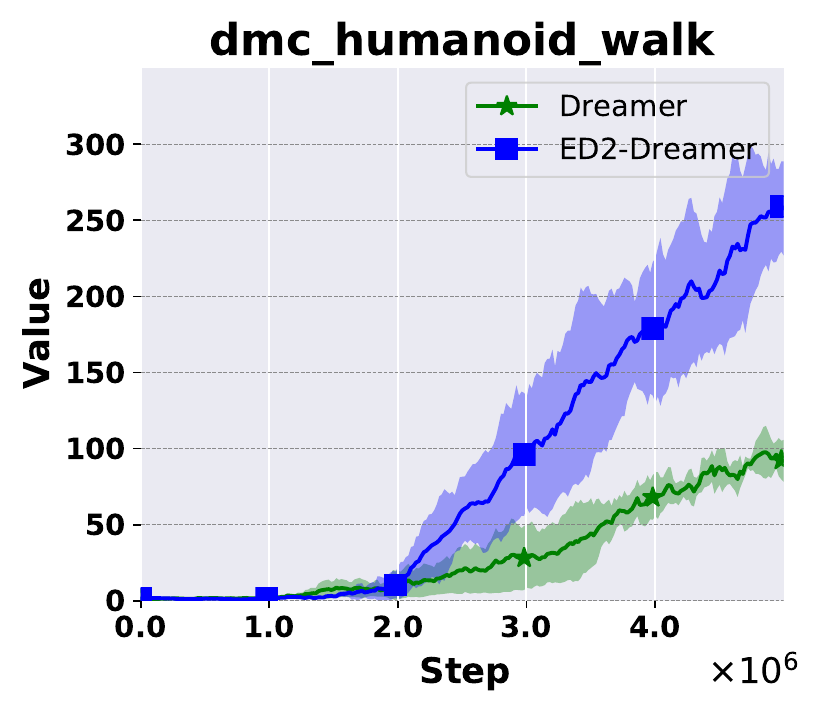}
\end{minipage}
\begin{minipage}[t]{0.3\linewidth}
\includegraphics[width=\linewidth]{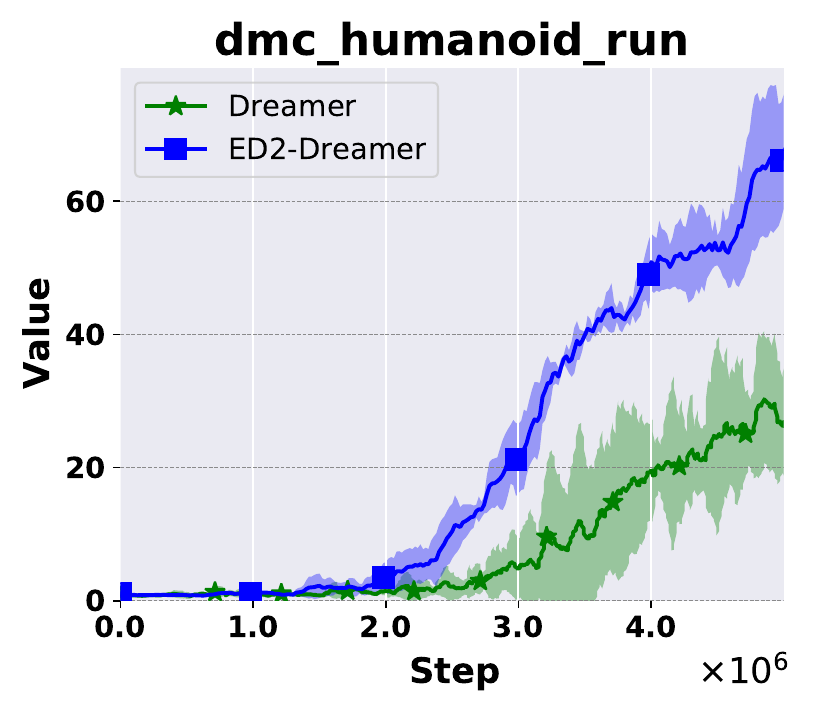}
\end{minipage}

\caption{Comparisons of \alg-Dreamer vs. Dreamer. The x- and y-axis represent the environment steps and the performance. The line and shaded area denote the mean value and standard deviation.}
\label{fig:alg-Performance}
\end{minipage}
\end{figure*}

\section{Experiments} \label{sec:exp}
% To evaluate the effectiveness of \alg, we conduct experiments for the following research questions:

% \noindent\textbf{RQ1~(Performance):} Can \alg improve the performance of existing MBRL algorithms?\\
% \noindent\textbf{RQ2~(Ablation):} What are the contributions of each component in \alg?\\
% \noindent\textbf{RQ3~(Accuracy):} Can \alg reduce model error when deployed in the MBRL learning process?\\
% \noindent\textbf{RQ4~(Subjective):}
% What is the performance of using subjective methods (HP) in SD2?
% Comparing the partition discovered by SD2 and the partition decided by human intuition, which is more reasonable in practical design?

\textbf{Baselines \& Benchmarks:}
For a comparative study, we take three state-of-the-art MBRL methods: MBPO \cite{DBLP:conf/nips/JannerFZL19}, Dreamer~\cite{DBLP:conf/iclr/HafnerLB020} and TDMPC~\cite{hansen2022temporal} as the baselines, and extend them with \alg as \alg-MBPO, \alg-Dreamer and \alg-TDMPC. To reduce implementation bias, we reuse the code and benchmarks from previous works: the DMControl Suite \cite{DBLP:journals/corr/abs-1801-00690} for Dreamer and TDMPC, Gym-Mujoco \cite{DBLP:journals/corr/BrockmanCPSSTZ16} for MBPO. To further validate the modeling capabilities of \alg for complex environments, we opt for DMC Remastered (DMCR)~\cite{grigsby2020measuring} which is a challenging extension of the DMC with randomly generated graphics emphasizing visual diversity. Furthermore, we also provide ablation studies to validate the effectiveness of each component of our \alg. All benchmark examples are shown in \cref{fig:benchmarks}.

\textbf{Clarification:}
% (1) The data required by the clustering-based method is included in the figures, which is tiny relative to the whole training process.
(1) Although the data required by the clustering-based method is tiny (less than 1\% of the policy training), we include it in the figures for a fair comparison.
(2) We take the clustering-based method as the main SD2 implementation (denoted as \alg-Methods) and discuss other SD2 methods in \cref{sec:RQ4}~(complete decomposition, human prior are denoted as CD, HP respectively).
% % The data required by the clustering process is included in the figures, which is not obvious since it is less than 1\% of the policy training.
% (2) Since clustering-based method is well scalable and better performed, we take it as the SD2 implementation in our experiments (denoted as \alg-Methods).
% (3) We also provide the result of CD in the appendix and discuss the HP method separately in Section \ref{sec:RQ4} since it is highly subjective.
% (3) Due to the space limit, we leave the result of MBPO/\alg-MBPO in the appendix.
% (5) CL methods struggle in the humanoid tasks since the high complexity, therefore we replace the image-based state to vector-based state in the clustering process, this will be further discussed in Section \ref{discussion}.
(3) If not specifically labeled, all experiments ran with five seeds.

\textbf{Hyperparameters:} We keep the experiment setting the same with original baselines. Specifically, for the dynamics models, we need to maintain the hidden size of each sub-dynamic model, as shown in \cref{tab:param}. For the clustering process, the hyperparameter $\eta$ describes the tightness of constraints on inter-group distance and intra-group distance. When $\eta = 0$, the clustering process stop condition is that the distance between the two most relative clusters is equal to the distance between these two clusters and others (this is a general condition in the most environment). In some environments, although $\eta = 0$ is a good choice, but the value can be further finetuned to obtain more reasonable clustering
results. 

\begin{table}[h!]
\caption{The hidden size and $\eta$ value for each environment.}
\centering
\renewcommand{\arraystretch}{1.2}
\begin{tabular}{lcc}
\hline
\textbf{ENVIRONMENT} & \textbf{HIDDEN SIZE} & $\boldsymbol{\eta}$ \\
\hline
HOPPER~(DeepMind) & 200 & 0 \\
WALKER~(DeepMind) & 200 & -0.06 \\
CHEETAH~(DeepMind) & 200 & -0.1 \\
HUMANOID~(DeepMind) & 200 & 0 \\
REACHER~(DeepMind) & 200 & 0 \\
FINGER~(DeepMind) & 200 & 0 \\
HALFCHEETAH~(Gym-Mujoco) & 150 & 0 \\
HOPPER~(Gym-Mujoco) & 200 & -0.3 \\
WALKER~(Gym-Mujoco) & 200 & -0.2 \\
ANT~(Gym-Mujoco) & 150 & -0.12 \\
\hline
\end{tabular}
\label{tab:param}
\end{table}

% \vfill
% \begin{minipage}[t]{0.24\linewidth}
% \includegraphics[width=\linewidth]{Hopper.pdf}
% \end{minipage}
% \begin{minipage}[t]{0.24\linewidth}
% \includegraphics[width=\linewidth]{HalfCheetah.pdf}
% \end{minipage}
% \begin{minipage}[t]{0.24\linewidth}
% \includegraphics[width=\linewidth]{Ant.pdf}
% \end{minipage}
% \begin{minipage}[t]{0.24\linewidth}
% \includegraphics[width=\linewidth]{Walker2d.pdf}
% \end{minipage}

% \caption{Comparisons of \alg-Dreamer vs. Dreamer on DMC tasks (first two rows), \alg-MBPO vs. MBPO on Gym-Mujoco tasks (last row). The x- and y-axis represent the training steps and performance. The line denotes the mean value and the shaded area represents the standard deviation.}

\subsection{Performance} \label{sec:RQ1}

\textbf{Image inputs experiments:} First we evaluated Dreamer and \alg-Dreamer on nine DMC tasks with image inputs. As shown in \cref{fig:alg-Performance},
\alg-Dreamer outperforms Dreamer on all tasks. This is because \alg establishes a more rational and accurate world model, which leads to more efficient policy training. Another finding is that, in tasks like cheetah\_run, and walker\_run, \alg-Dreamer achieves lower variance, demonstrating that \alg can also lead to a more stable training process. 
Furthermore, Dreamer fails to achieve good performance in difficult tasks such as humanoid domain. In contrast, \alg-Dreamer improves the performance significantly, which indicating the superiority of \alg in complex tasks. The \alg framework can simply adopt to different types of MBRL algorithms and input types. In the evaluation of sampling efficiency, we chose the image-based DMControl 200k benchmark and adopted the of state-of-the-art algorithm TDMPC planning based in this benchmark as the backbone, and \cref{tab:TDMPC-table} shows the experimental results. \alg-TDMPC shows a steady improvement in all environments over baseline, proving the validity of the ED2 world model structure across different types of world model usage.

\begin{figure}[ht]
\centering
\begin{minipage}{0.46\linewidth}
\includegraphics[width=\linewidth]{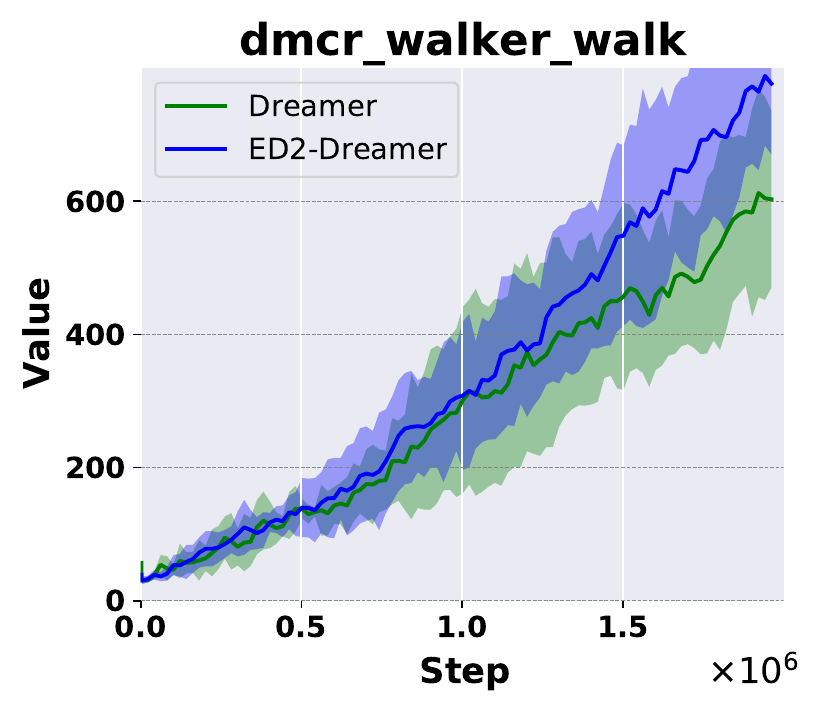}
\end{minipage}
\begin{minipage}{0.46\linewidth}
\includegraphics[width=\linewidth]{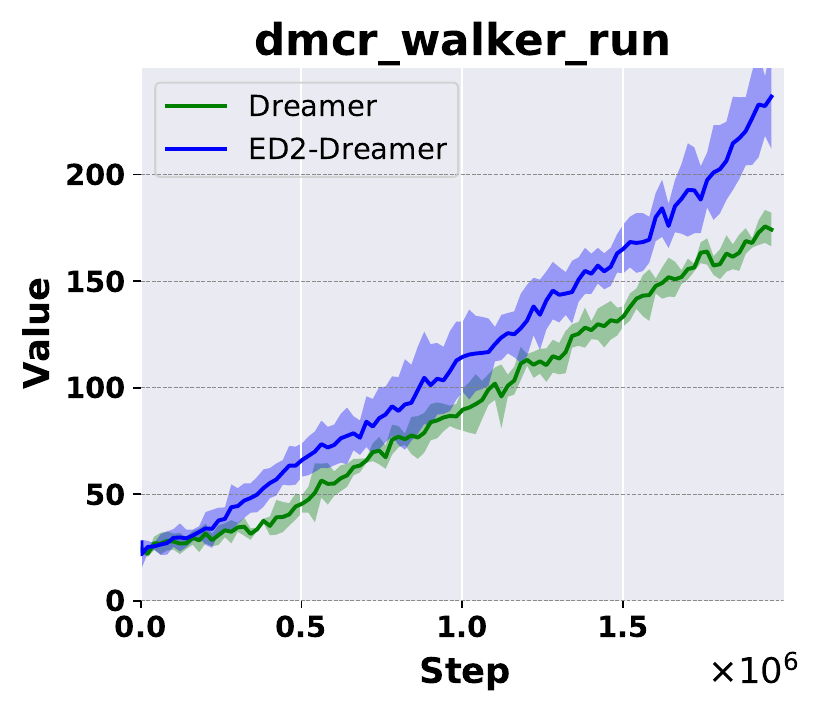}
\end{minipage}
\caption{Comparisons of \alg-Dreamer in DMC Remastered benchmark. Walker needs to learn to walk or run with randomly generated background.}
\label{fig:dmcr}
\vspace{-15pt}
\end{figure}

\begin{figure*}[htbp]
\centering
\begin{minipage}[t]{\linewidth}
\centering
\begin{minipage}[t]{0.24\linewidth}
\centering
\includegraphics[width=\linewidth]{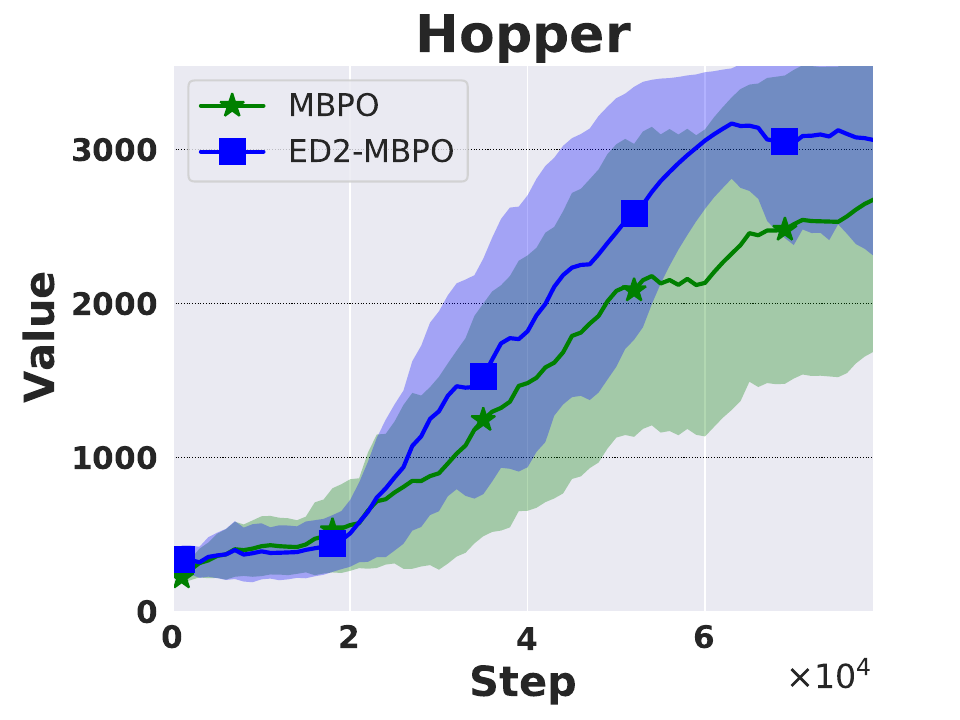}
\end{minipage}
\begin{minipage}[t]{0.24\linewidth}  \includegraphics[width=\linewidth]{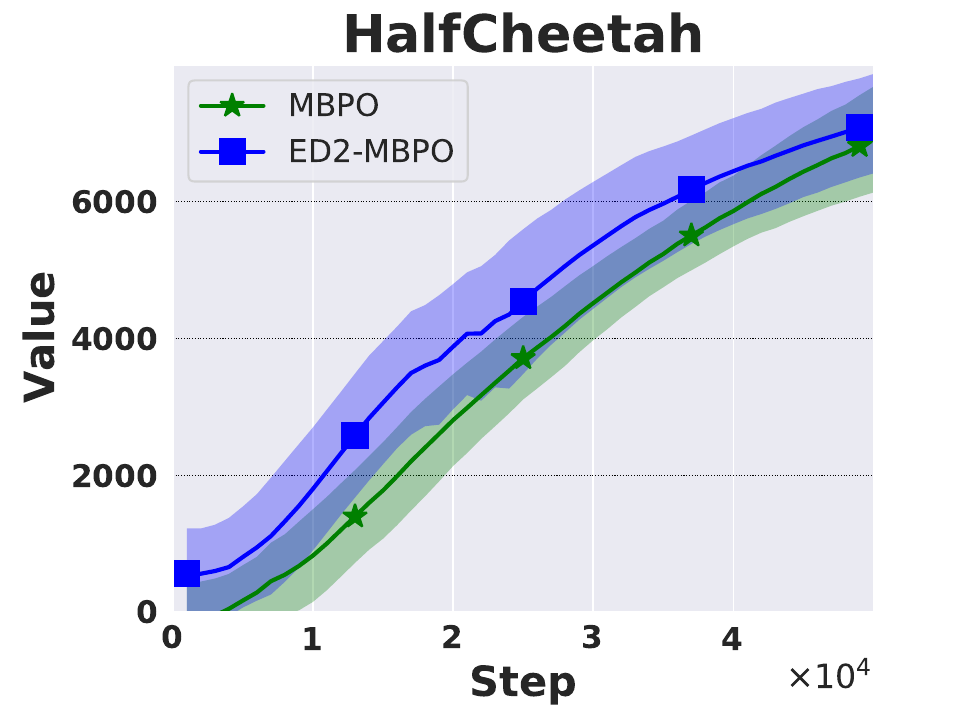}
\end{minipage}
\begin{minipage}[t]{0.24\linewidth}
\includegraphics[width=\linewidth]{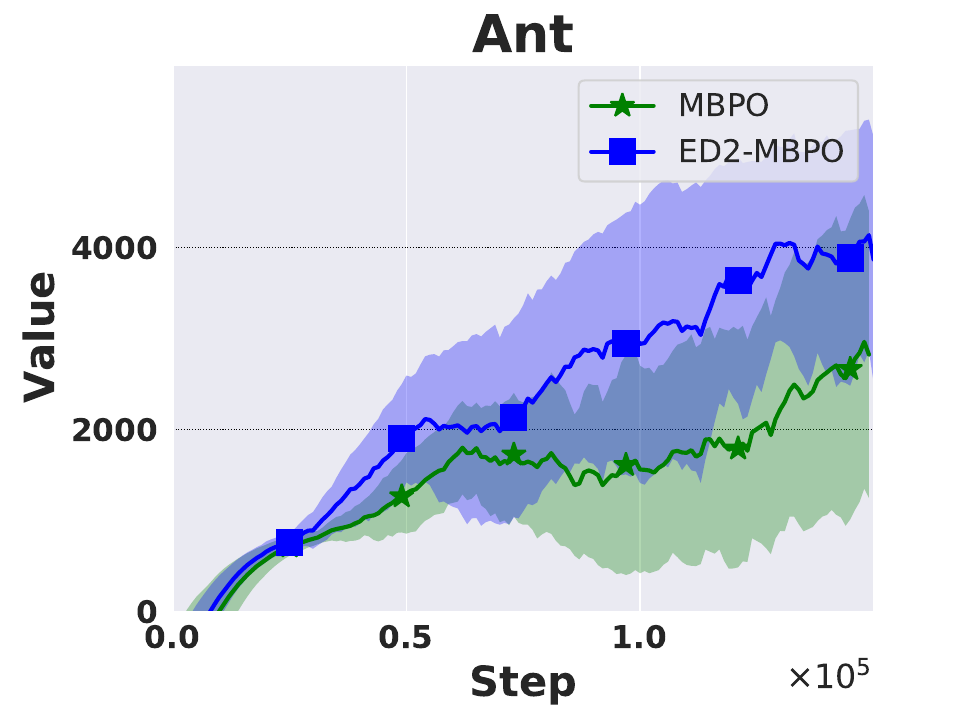}
\end{minipage}
\begin{minipage}[t]{0.24\linewidth}
\includegraphics[width=\linewidth]{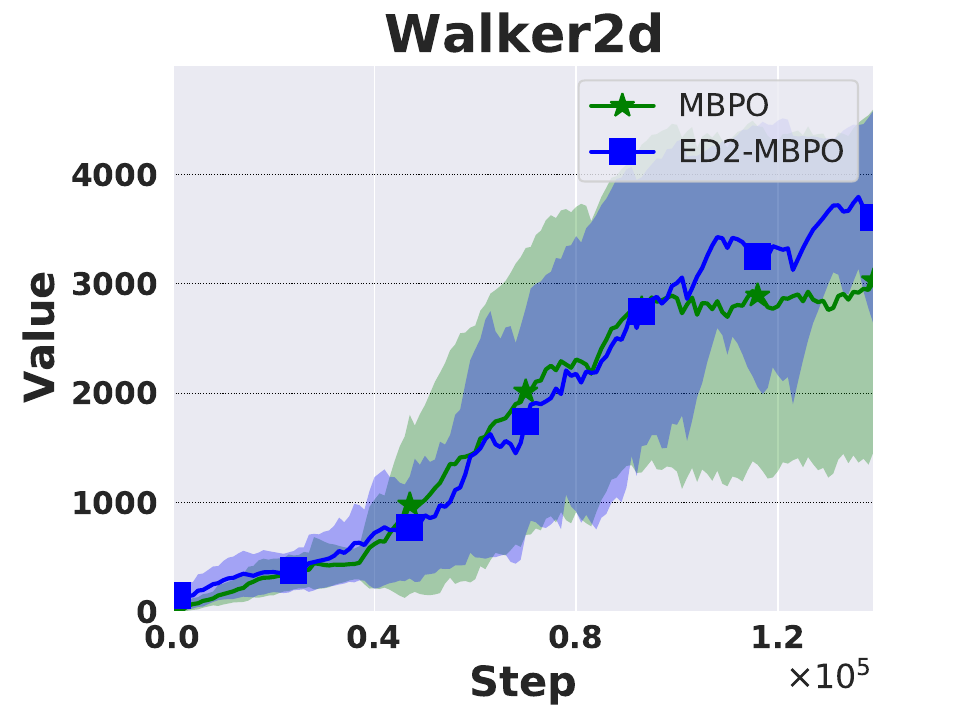}
\end{minipage}
\caption{Performance comparisons between \alg-MBPO and MBPO. The line and shaded area denote the mean value and standard deviation.}
\label{fig:mbpo_performance}
\end{minipage}
\end{figure*}

\textbf{Proprioceptive state inputs experiments:}  Also, we provide the comparison of MBPO and \alg-MBPO in the \cref{fig:mbpo_performance}, which proves that \alg significantly boost the performance of MBPO.

\textbf{Image inputs experiments with complicated graphical factors:}  Furthermore, we validate ED2's ability to model environmental dynamics in more complex scenarios. In DMCR, the environment is subject to multiple visual influences such as loor texture, background, robot body color, etc., which poses great difficulties for dynamic modeling. In this case, \alg-Dreamer is able to produce a stable performance improvement to the baseline. This shows that \alg can effectively learn the shared knowledge of action space decomposition, even if the environmental observation is exceptionally complex.

\begin{table*}[h]
\caption{Performance of \alg-TDMPC and \alg on the sample-efficiency image-based DMControl 200k benchmark. The better results are highlighted in bold. All experiments runs with 3 seeds.}
\label{tab:TDMPC-table}
\begin{center}
\resizebox{0.99\linewidth}{!}{
\begin{tabular}{l|ccccccc}
\hline
\textit{200k env. steps} & Hopper-Hop & Cup-Catch & Reacher-Hard & Cheetah-Run & Finger-Spin & Walker-Walk & Walker-Run \\
\hline
ED2-TDMPC & \textbf{95.3 $\pm$ 49.8} & \textbf{761.2 $\pm$ 152.2} & \textbf{94.8 $\pm$ 53.3} & \textbf{340.4 $\pm$ 57.8} & \textbf{976.6 $\pm$ 7.3} & \textbf{884 $\pm$ 25.7} & \textbf{406 $\pm$ 17.6} \\
TDMPC     & 60.0 $\pm$ 37.3 & 749.5 $\pm$ 177.5 & 34.6 $\pm$ 25.4 & 263.8 $\pm$ 49.1 & 875.0 $\pm$ 109.0 & 867 $\pm$ 21.6 & 366 $\pm$ 7.8 \\
\hline
\end{tabular}}
\end{center}
\end{table*}

\subsection{Ablation Studies}\label{sec:RQ2}
In this section, we verify the effectiveness of D2P and SD2 and investigate the contribution of each component to performance improvement.
We can summarize the improvements into three parts: \textbf{multiple kernels, decomposing prediction, reasonable partition}.
There is a progressive dependence between these three parts: the decomposing prediction depends on the existence of the multiple kernels and the reasonable partition depends on the decomposing prediction. Therefore, we design an incremental experiment for the validation. 

First, we employ the \alg-Dreamer-ensemble, which maintains the multiple-kernel structure but without dynamics decomposing (i.e. all kernels input with action $a$ rather than sub-action $a^{G_i}$). In MBRL, the model ensemble is generally implemented by setting different initial parameters and sampling different training data from the same dataset. Therefore, the model ensemble is totally different from the kernel ensemble used by \alg. Furthermore, the ensemble model proposes to estimate the prediction uncertainty, which is orthogonal to \alg. Both methods can be utilized to decrease model error. We investigate the contribution of multiple kernels by comparing \alg-Dreamer-ensemble with baselines.
Second, we employ the \alg-Dreamer-random, which maintains the D2P structure and obtains partition randomly. We investigate the contribution of decomposing prediction by comparing \alg-Dreamer-random with \alg-Dreamer-ensemble.
Finally, we investigate the contribution of reasonable partition by comparing \alg-Dreamer with \alg-Dreamer-random.

\begin{figure*}[ht]
\centering
\begin{minipage}{0.24\linewidth}
\includegraphics[width=\linewidth]{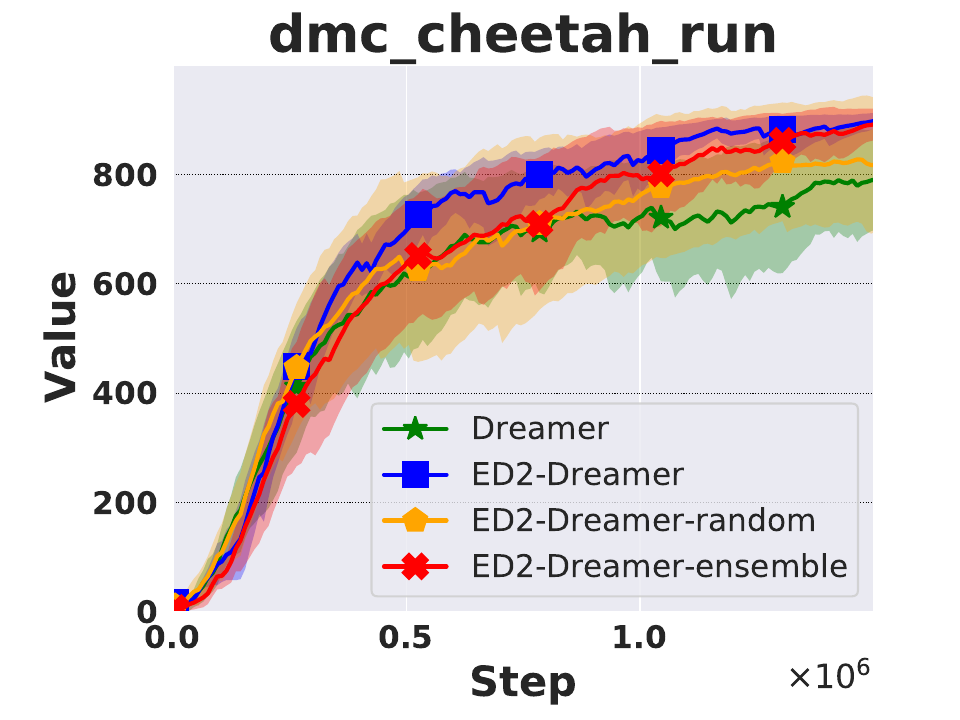}
\end{minipage}
\begin{minipage}{0.24\linewidth}
\includegraphics[width=\linewidth]{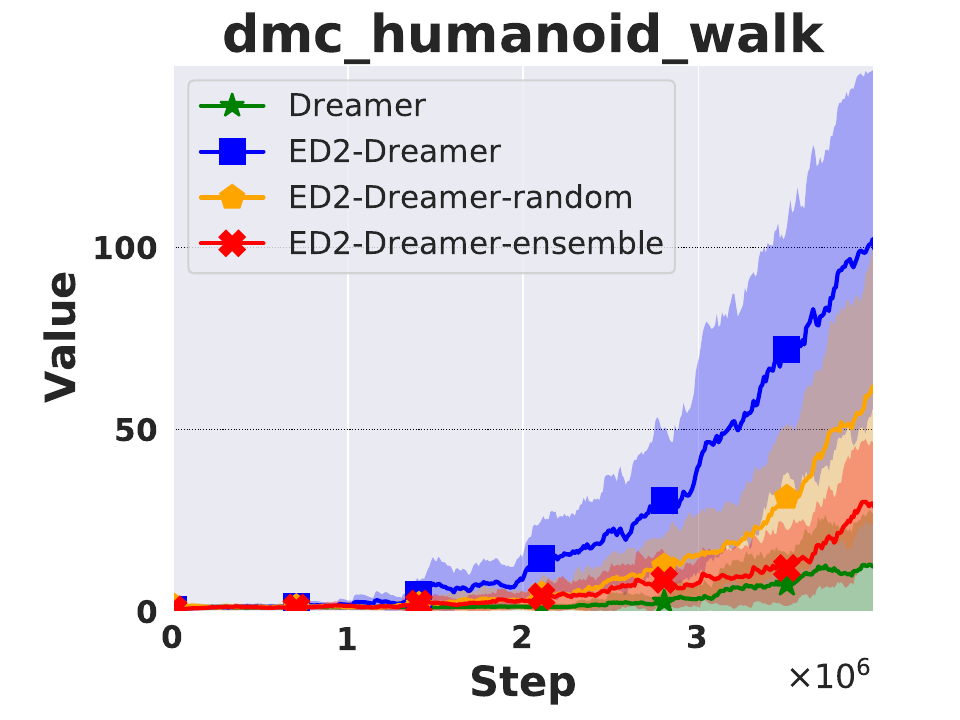}
\end{minipage}
\begin{minipage}{0.24\linewidth}
\includegraphics[width=\linewidth]{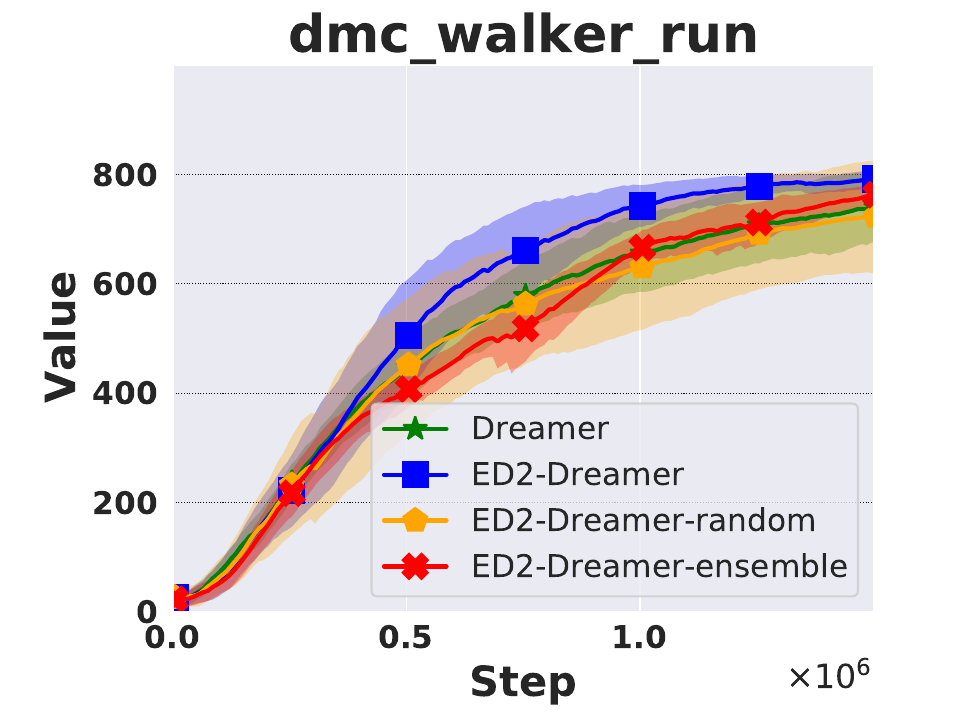}
\end{minipage}
\begin{minipage}{0.24\linewidth}
\includegraphics[width=\linewidth]{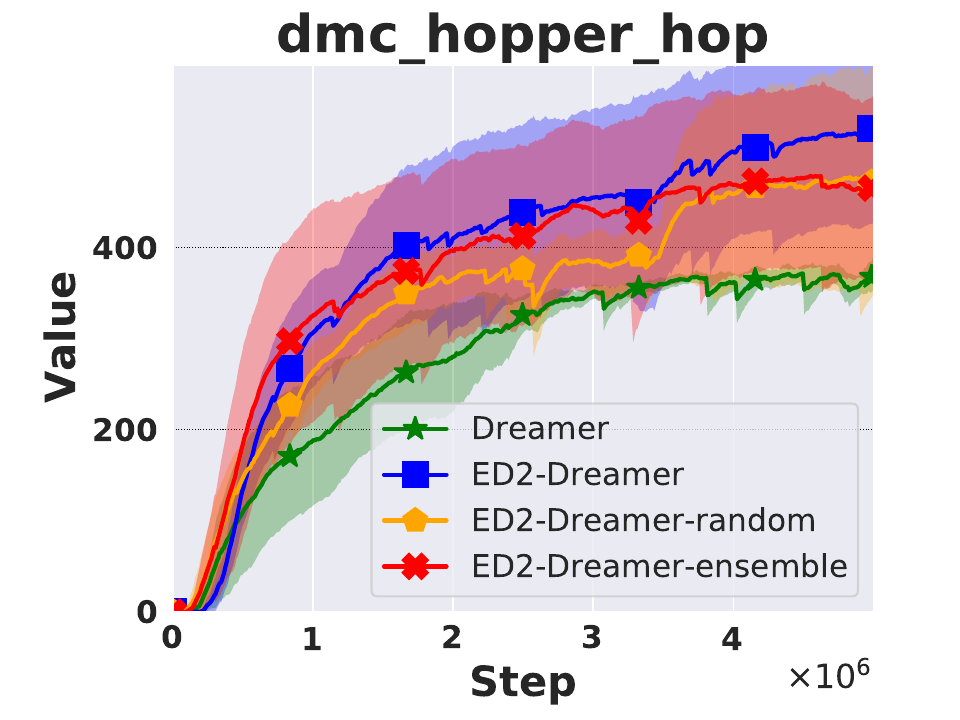}
\end{minipage}
\caption{Performance comparisons of components ablation experiments.}
\label{fig:ablation_od}
\end{figure*}

% \begin{figure}
% \centering
% % \vspace{-0.8cm}
% \begin{minipage}[t]{\linewidth}
% \centering
% \begin{minipage}[t]{0.7\linewidth}
% \centering
% \includegraphics[width=\linewidth]{figures/ablation_dmc_cheetah_run.pdf}
% \end{minipage}
% \begin{minipage}[t]{0.7\linewidth}  \includegraphics[width=\linewidth]{figures/ablation_hopper.pdf}
% \end{minipage}
% \vspace{-15pt}
% \caption{Extension of RSSM with D2P.}
% \label{fig:ablation_state}
% \end{minipage}
% \end{figure}

\cref{fig:ablation_od} shows that
\alg-Dreamer-ensemble outperform Dreamer on humanoid\_walk and cheetah\_run tasks, indicating that multiple kernels help the policy training on some tasks.
\alg-Dreamer-random outperforms \alg-Dreamer-ensemble on humanoid\_walk task, but not in other tasks (even perform worse in cheetah\_run and hopper\_hop). This is due to the different modeling difficulty of tasks: the tasks except humanoid\_walk are relatively simple and can be modeled without D2P directly (but in a sub-optimal way). The modeling process of these tasks can be aided by a reasonable partition but damaged by a random partition. The humanoid\_walk is challenging and cannot be modeled directly, therefore decomposing prediction (D2P) is most critical and performance can be boosted even with a random decomposing prediction.
Finally, \alg-Dreamer outperforms \alg-Dreamer-random on all tasks, which indicates that a reasonable partition (SD2) is critical in dynamics modeling and D2P can not contribute significantly to the modeling process without a reasonable partition.

% \begin{figure}[t]
% \centering
% \begin{minipage}[t]{\linewidth}
% \centering
% \begin{minipage}[t]{0.24\linewidth}
% \includegraphics[width=\linewidth]{ablation_dmc_cheetah_run.pdf}
% \end{minipage}
% \begin{minipage}[t]{0.24\linewidth}  \includegraphics[width=\linewidth]{ablation_dmc_humanoid_walk.pdf}
% \end{minipage}
% \begin{minipage}[t]{0.24\linewidth}
% \includegraphics[width=\linewidth]{ablation_dmc_walker_run.pdf}
% \end{minipage}
% \begin{minipage}[t]{0.24\linewidth}
% \includegraphics[width=\linewidth]{ablation_dmc_walker_walk.pdf}
% \end{minipage}
% \caption{Performance comparisons of components ablation experiments.}
% \label{graph:ablation}
% \end{minipage}
% \end{figure}

\subsection{Model Error} \label{sec:RQ3}
\begin{figure*}[ht]
\begin{minipage}[t]{\linewidth}
\centering
\includegraphics[width=0.24\linewidth]{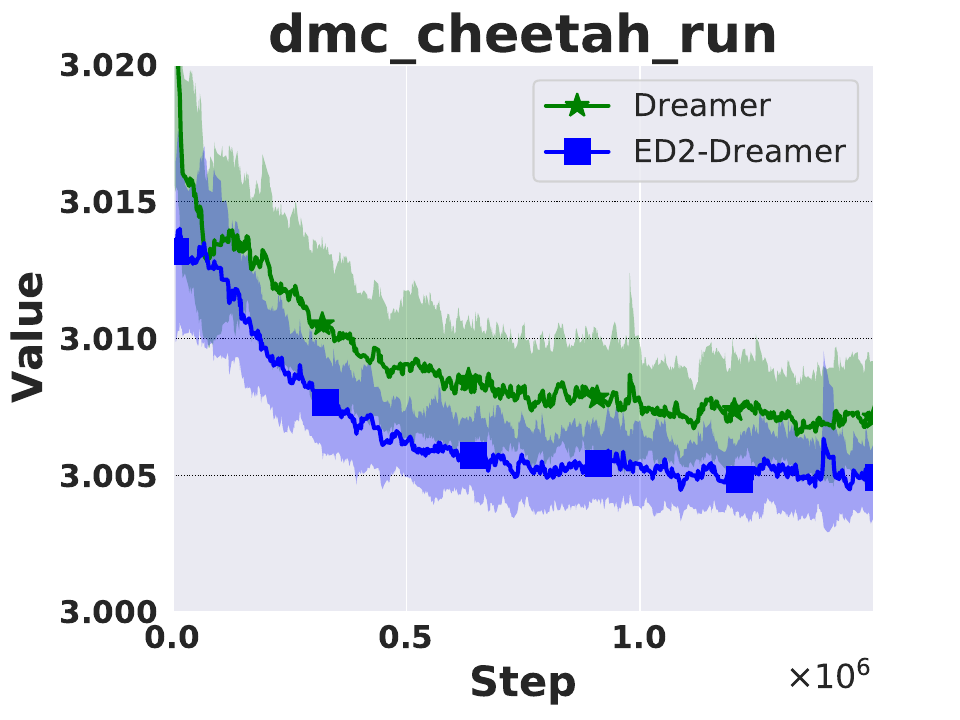}
\includegraphics[width=0.24\linewidth]{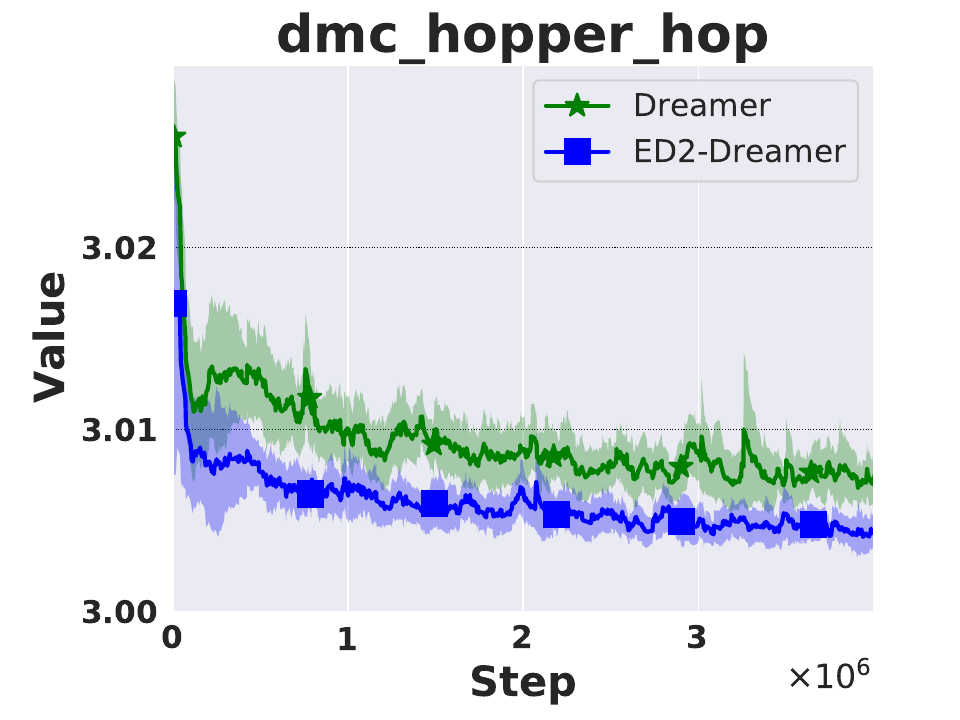}    
\includegraphics[width=0.24\linewidth]{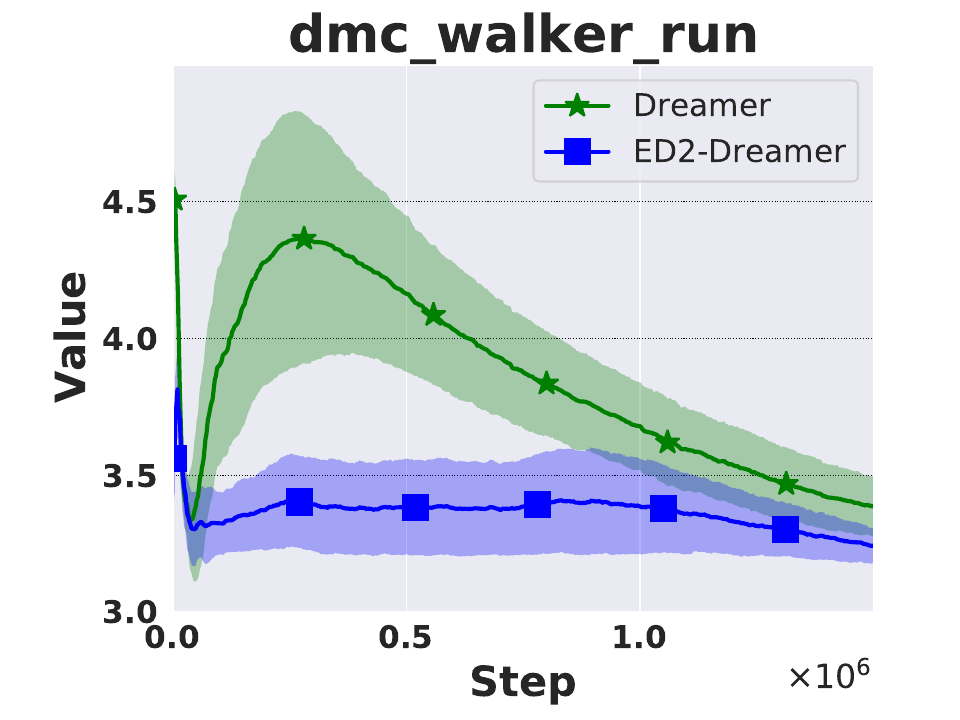}
\includegraphics[width=0.24\linewidth]{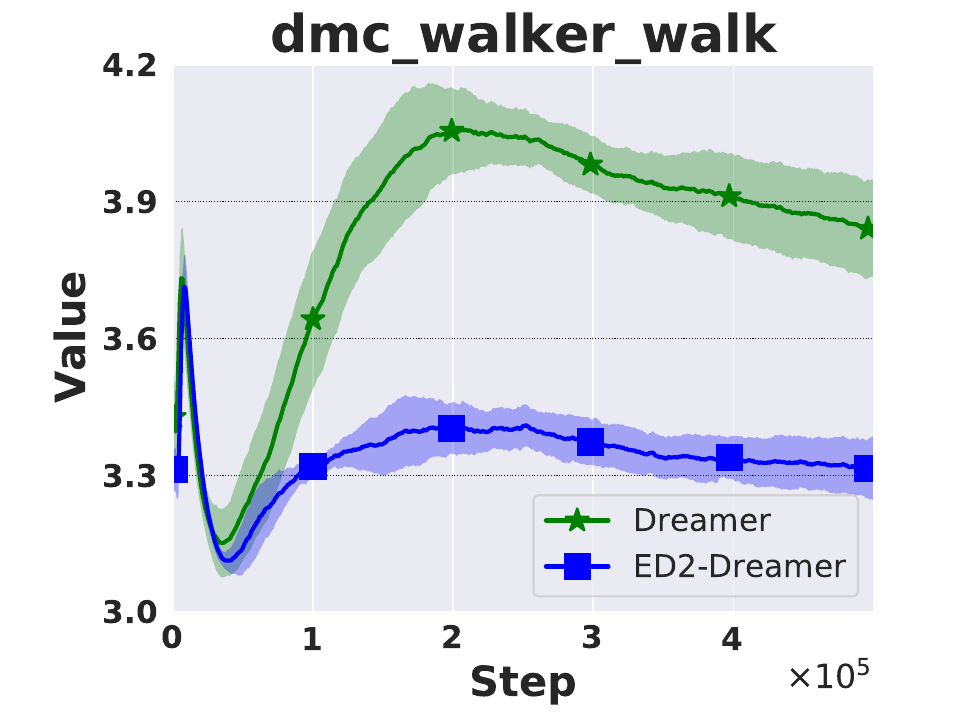}
\includegraphics[width=0.24\linewidth]{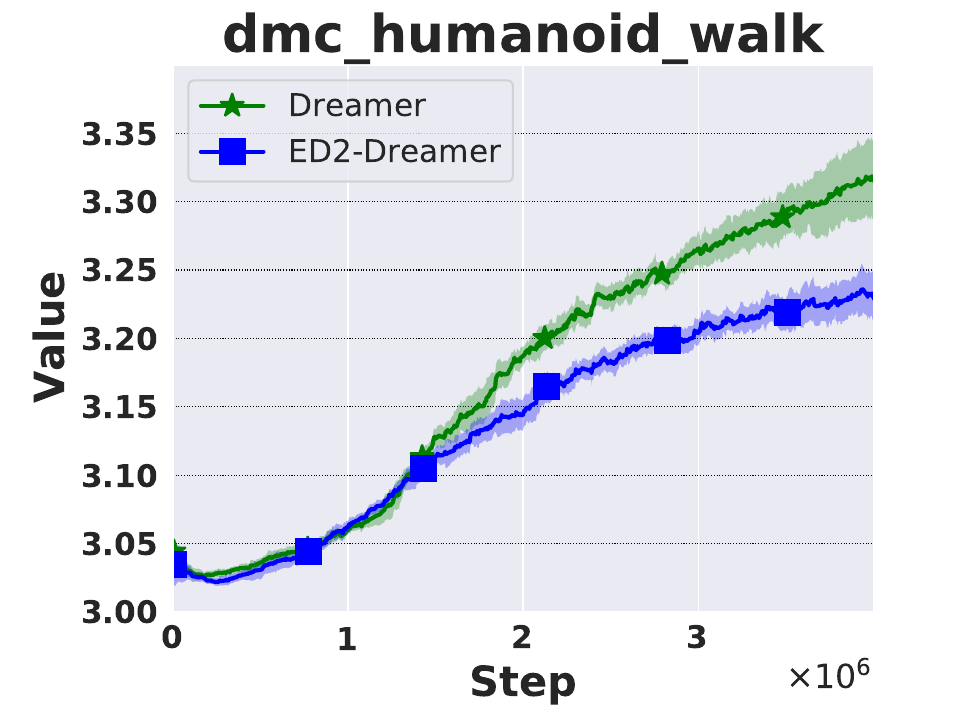}
\includegraphics[width=0.24\linewidth]{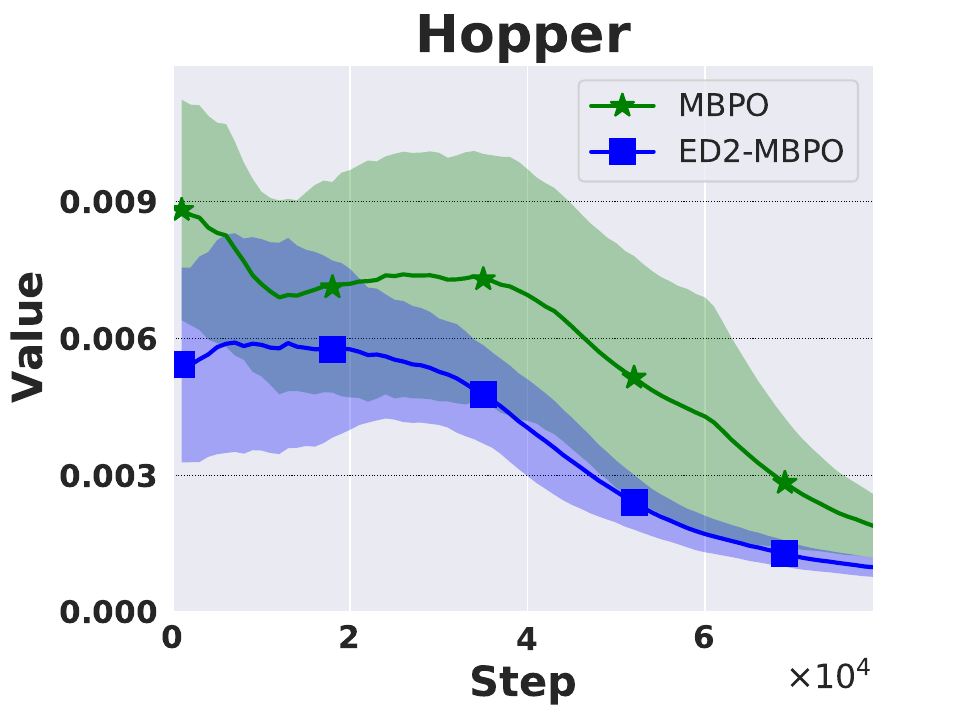}
\includegraphics[width=0.24\linewidth]{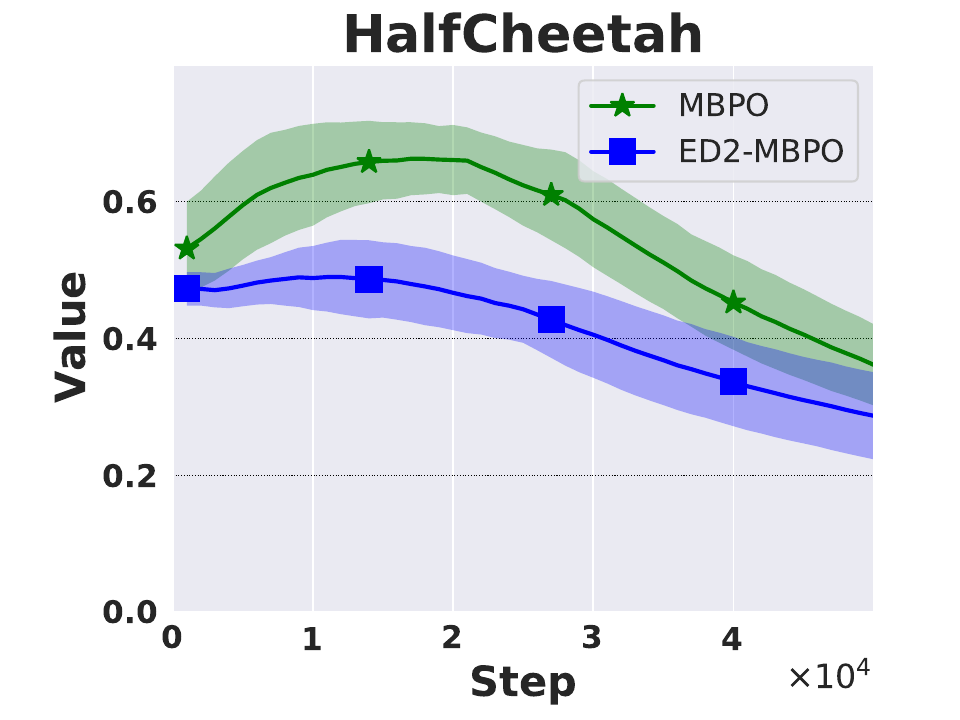}
\includegraphics[width=0.24\linewidth]{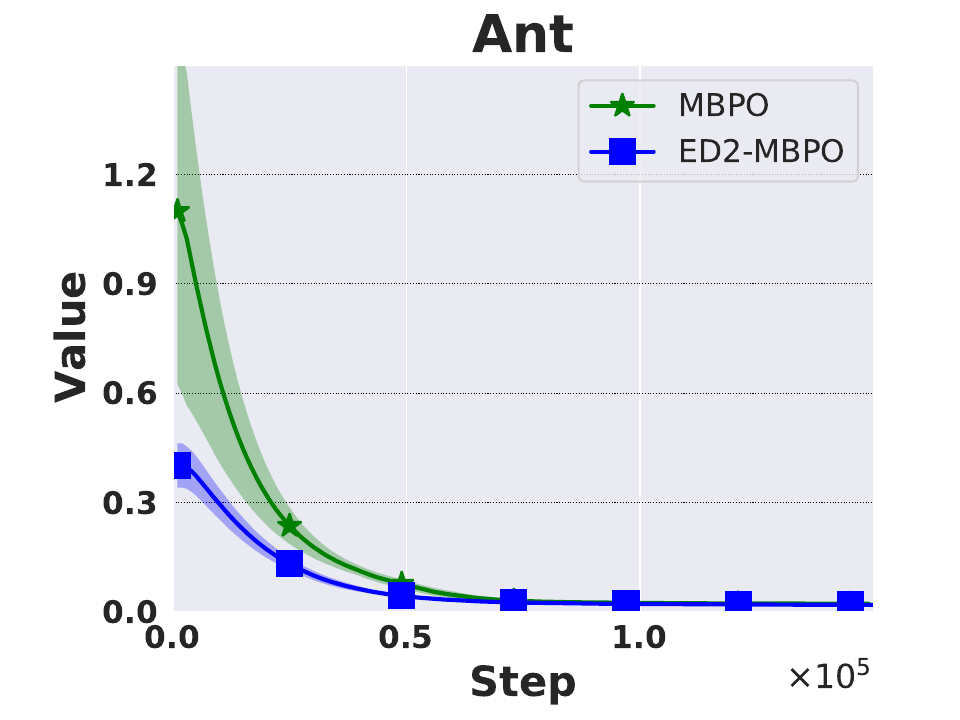}

\caption{The model error comparison of \alg-methods and baselines. The x- and y-axis denote the training steps and the model error (MSE-Loss in MBPO / \alg-MBPO and KL-Divergence in Dreamer / \alg-Dreamer), respectively.}
%\caption{The model error (KL-Divergence) comparison of \alg-Dreamer and Dreamer.} 
\label{fig:model error}
\end{minipage}
\end{figure*}

% \begin{figure*}[h]
% \centering
% \begin{minipage}[t]{\linewidth}
% \centering
% \begin{minipage}[t]{0.24\linewidth}
% \centering
% \includegraphics[width=\linewidth]{Hopper_loss.pdf}
% \end{minipage}
% \begin{minipage}[t]{0.24\linewidth} \includegraphics[width=\linewidth]{HalfCheetah_loss.pdf}
% \end{minipage}
% \begin{minipage}[t]{0.24\linewidth}
% \includegraphics[width=\linewidth]{Ant_loss.pdf}
% \end{minipage}
% \begin{minipage}[t]{0.24\linewidth}
% \includegraphics[width=\linewidth]{Walker2d_loss.pdf}
% \end{minipage}
% \caption{The model error comparisons between \alg-MBPO and MBPO.}
% \label{fig:MBPO model error}
% \end{minipage}
% \end{figure*}

In this section, we further investigate whether the model error is reduced when combined with \alg. We conducted an environment-modeling experiment on the same dataset (which is collected in the MBRL training process) and recorded the model error.
Since the policy keeps update in the MBRL training process, the MBRL dataset also changes with the updated policy.
For example, in the MBRL training on the Cheetah task, the model is first trained with the data like \textit{Rolling on the ground} and finally trained with the data like \textit{Running with high speed}.
To simulate the MBRL training process, we implement our dataset by slowly expanding it from 0 to all according to the data generation time.
% To make our dataset more similar to the dataset in MBRL training, we slowly expanded our dataset from 0 to all according to the data generation time.
This setting can also help to investigate the generalization ability of the model on unfamiliar data (i.e., the data generated by the updated policy).
We list the results in \cref{fig:model error}. \cref{fig:model error} shows that \alg-Dreamer has a significantly lower model error in all tasks compared with Dreamer. \alg-Dreamer can also achieve a more stable world model training (i.e. with low variance) and on significantly increasing model error humanoid\_walk and walker\_run tasks. The comparison of model error of MBPO and \alg-MBPO in the \cref{fig:model error} also proves that \alg can reduce the model errors when combined with MBRL methods.
% Both MBRL algorithms can obtain a more accurate world model by combining with \alg, which indicates that \alg is a general framework for existing MBRL algorithms.
We hypothesize that \alg produces a reasonable network structure; as the dataset grows, \alg-methods can generalize to the new data better. This property is significant in MBRL since the stable and accurate model prediction is critical for policy training.

\subsection{D2P Comparison} \label{sec:D2P_comparison}

\begin{figure*}[ht]
\centering
\begin{minipage}[b]{0.26\linewidth}
\centering
\includegraphics[width=\linewidth]{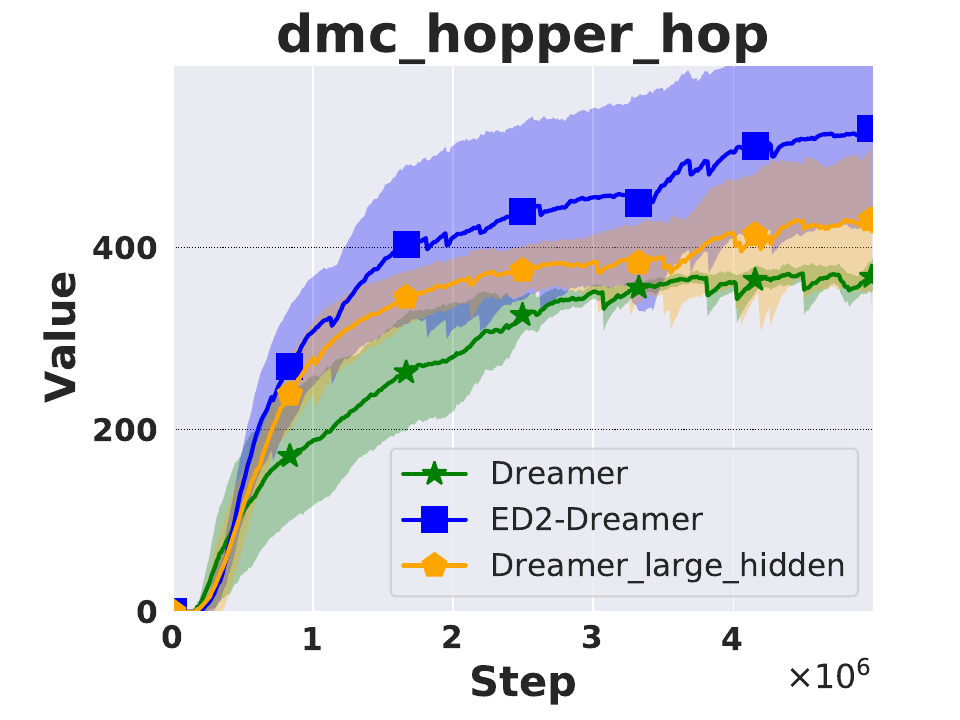}
\caption{The performance of Dreamer under bigger parameter size.}
\label{fig:large_hidden}
\end{minipage}
\hfill % 这会在两个minipage之间添加一些空间
\begin{minipage}[b]{0.7\linewidth}
\centering
\includegraphics[width=0.32\linewidth]{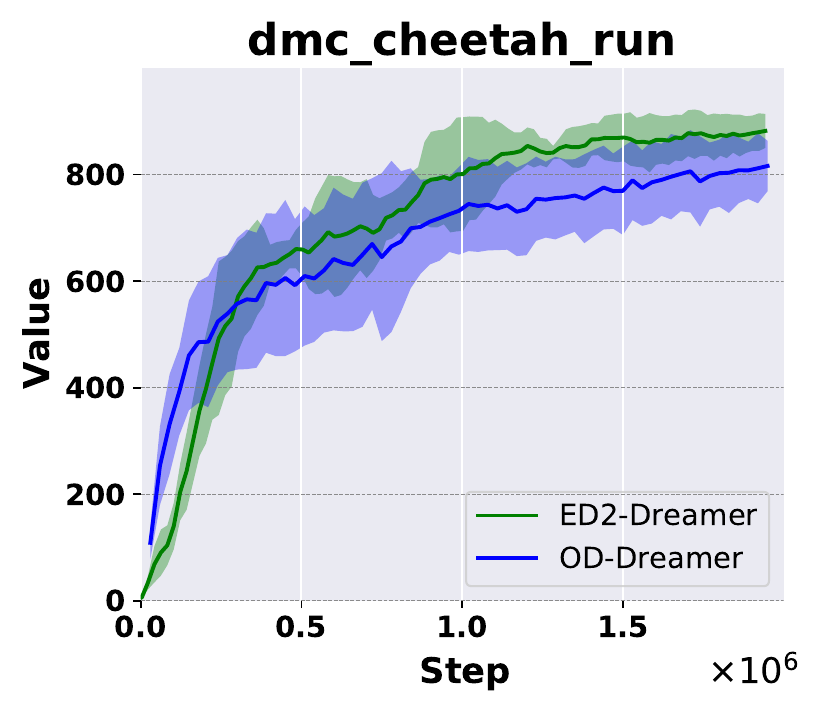}
\includegraphics[width=0.32\linewidth]{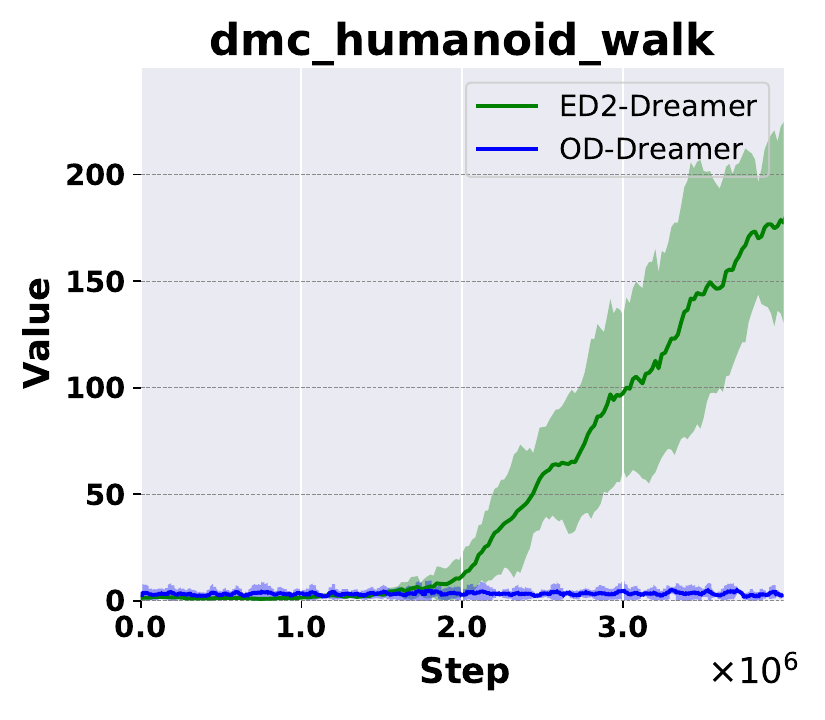}
\includegraphics[width=0.32\linewidth]{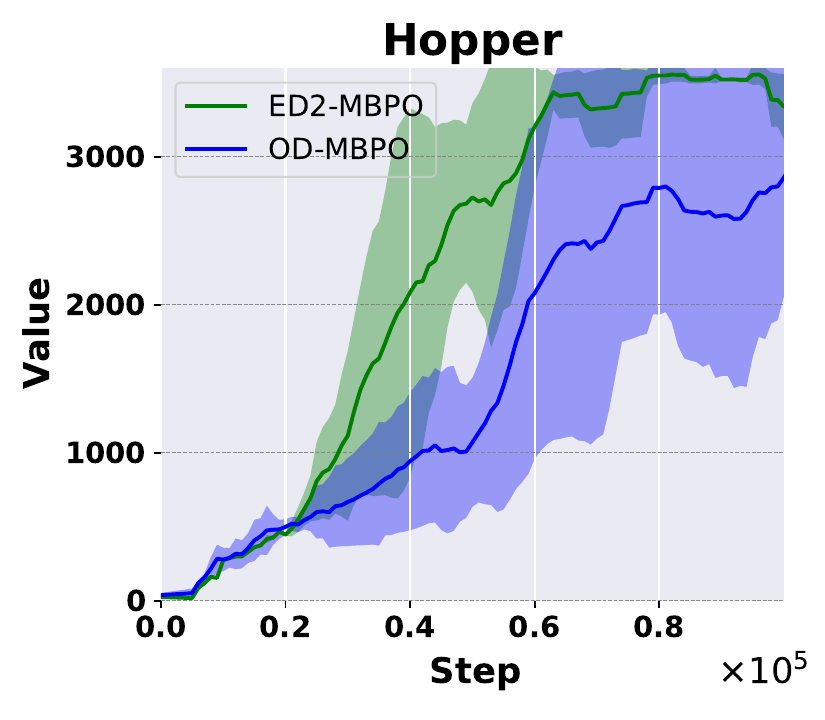}
\caption{The comparison of action decomposition~(\alg) and observation decomposition manner~(OD).}
\label{fig:od_dreamer}
\end{minipage}
\end{figure*}

The kernel ensemble used by \alg in the RSSM structure slightly increases the number of parameters of the Dreamer. For fair comparison, we provide the result of Dreamer method under bigger hidden size (which keeps the similar parameter size as \alg-Dreamer). As shown in \cref{fig:large_hidden}, increasing the size of parameters almost can not improve the performance.

Then, we compare the performance of a decomposition of the observation space at both image input and low-dimensional proprioceptive state input environments. As for image observation, it is difficult to decompose raw pixel inputs into sub-dynamics with explicit semantic information, whereas action space decomposition is unaffected. We modified the core part of the MWM~\cite{seo2023masked} method, which also based on Dreamer, to split the pixel input into patches, encode them using the vision transformer, and adaptively learn the observation decomposition dynamics (OD), called OD-Dreamer. We also implemented the observation decomposition version of the MBPO for low-dimensional state input, called OD-MBPO. OD-MBPO used different dynamics models to predict the angle and velocity of the agent respectively.
As shown in \cref{fig:od_dreamer}, the action decomposition mechanism used by \alg is superior to the OD methods, especially in the complex humanoid domain. The same result remains in the low-dimensional state inputs, because some uncontrollable and dynamically irrelevant factors are meaningless for decomposition. \alg decompose the dynamics from the root, i.e., the action space.

% \begin{figure*}[ht]
% \centering
% \begin{minipage}{0.24\linewidth}
% \includegraphics[width=\linewidth]{dmc_hopper_hop_bigger_image.pdf}
% \end{minipage}
% \begin{minipage}{0.24\linewidth}
% \includegraphics[width=\linewidth]{figures/dreamerv2/ablation_dmc_cheetah_run.pdf}
% \end{minipage}
% \begin{minipage}{0.24\linewidth}
% \includegraphics[width=\linewidth]{figures/dreamerv2/ablation_dmc_humanoid_walk.pdf}
% \end{minipage}
% \begin{minipage}{0.24\linewidth}
% \includegraphics[width=\linewidth]{figures/dreamerv2/ablation_dmc_humanoid_walk.pdf}
% \end{minipage}
% \caption{Performance comparisons of components ablation experiments.}
% \label{fig:action_decomposition}
% \end{figure*}

\subsection{SD2 Comparison} \label{sec:RQ4}

\begin{figure*}[t]
\centering
\includegraphics[width=0.95\linewidth]{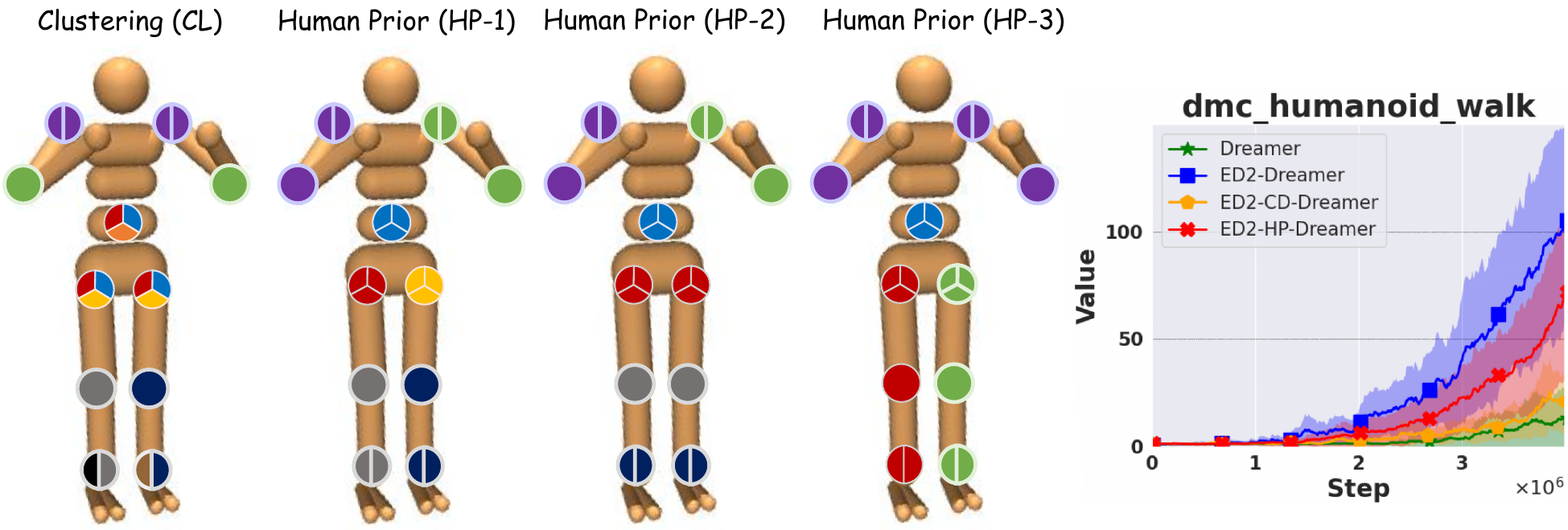}
\caption{The performance comparison of Dreamer, \alg-Dreamer, \alg-CD-Dreamer and \alg-HP-Dreamer. We provide the sub-dynamics visualization in the left four figures. Each circle correspond to a joint. 
A joint contains multiple action dimensions when the corresponding circle is separated into multiple parts. We mark the action dimensions in the same sub-dynamics with the same color.
}
\label{fig:CL_vs_HP}
\end{figure*}

\begin{figure*}[h]
\centering
\includegraphics[width=0.8\linewidth]{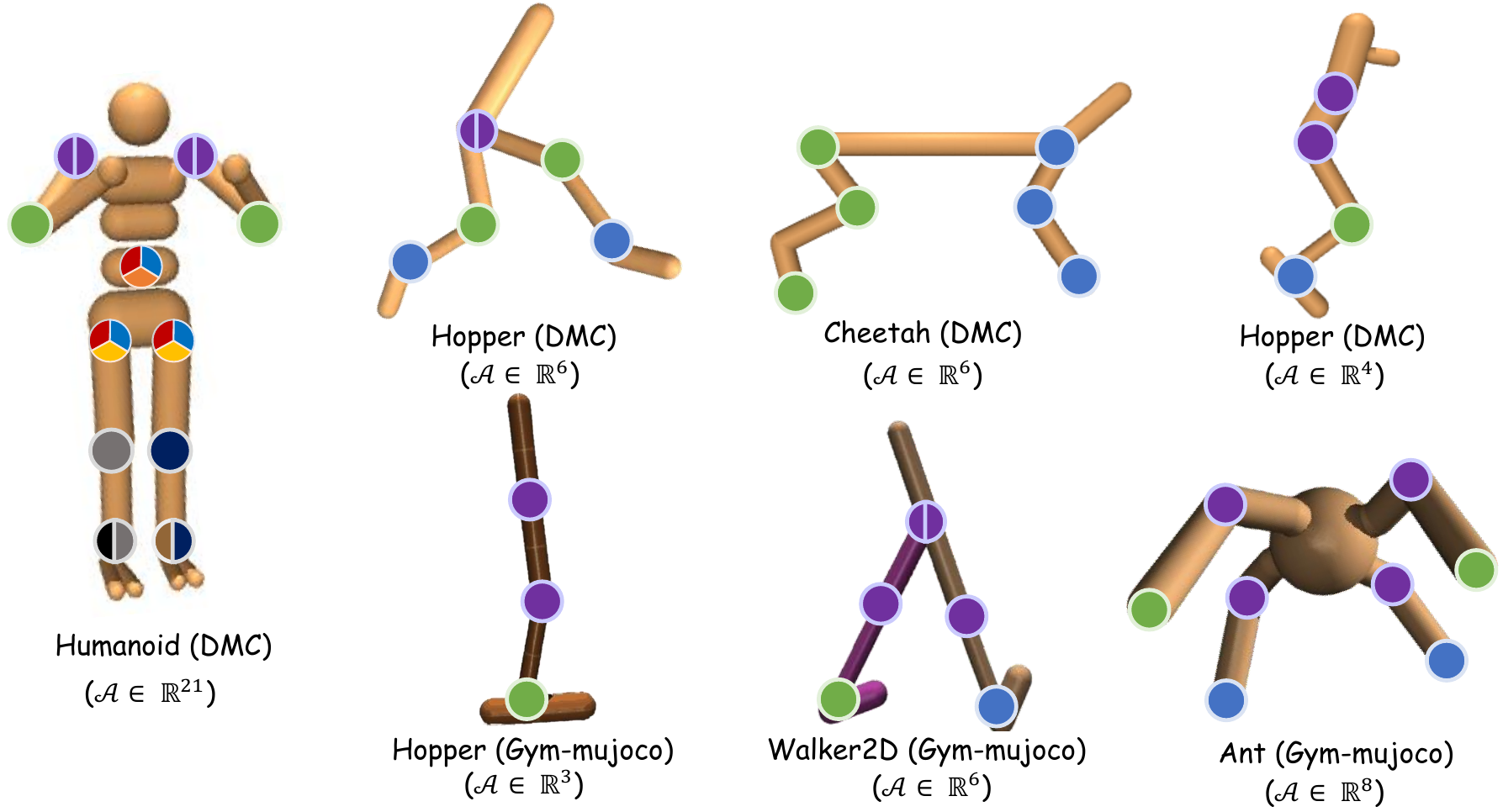}
\caption{The visualization of final partition of environments in DeepMind Control Suite and Gym-Mujoco.
}
\label{fig:decompose_result}
\end{figure*}

In this section, we compare the performance of three proposed SD2 methods. 
% we select humanoid\_walk as our environment and compare three SD2 methods on it.
We list the decomposition obtained by the clustering-based method and human prior on humanoid\_walk task for the illustrating purpose and provide the corresponding performance comparison results in \cref{fig:CL_vs_HP}.

% As show in the appendix, different SD2 methods achieves similar or identical results in simple tasks. Therefore, we select the complex environment humanoid\_walk here for the further investigation.
% We list the decomposition obtained by CL and designed by HP in Figure \ref{human_prior} and provide the performance comparison experiment for the investigation.

As shown in \cref{fig:CL_vs_HP}, human prior can generate different partitions from different task understandings and we average their performance as the final result. The experiment shows that all SD2 methods help policy learning. The clustering-based method performs best, and baseline Dreamer performs worst in the comparison. For the complete decomposition, it performs poorly under humanoid\_walk, which implies that humanoid\_walk contains many inner action dimension correlations, and simply complete decomposition heavily breaks this correlation, thus hindering the final performance. Compared to complete decomposition, human prior maintains more action dimension correlations by leveraging the human prior knowledge, leading to better performance. However, the correlations maintained by human prior might be false or incomplete due to human limited understanding of tasks. 
Compare to human prior, 
the clustering-based method automatically decomposes the action space according to the clustering criterion, which decomposes the action space better in a mathematical way. 
For example, human prior aggregates $\{right\_hip\_x, right\_hip\_y, right\_hip\_z\}$ ($x,y,z$ denote the rotation direction) together, and the clustering-based method aggregates $\{abdomen\_x, right\_hip\_x, left\_hip\_x\}$ together. Although the action dimensions from human prior sub-dynamics affect the same joint $left\_hip$, they rotate in different directions and play a different role in the dynamics. In contrast, the sub-dynamics discovered by the clustering-based method aggregate the action dimensions that affect the x-direction rotation together. It maintains stronger correlations and helps the world model fitting the movement on x-direction better. Therefore, it performs better than the human prior on this task. We list CL results of all environments in the \cref{fig:decompose_result}. The results show that most discovered partitions can be obtained with CL, which retains the correlation between similar affect action dimensions. Therefore, the clustering-based method discovers the closer correlations between action dimensions, which is not limited by the human intuitive.

\section{Conclusion}
In this paper, we propose a novel world model construction framework: Environment Dynamics Decomposition (\alg), which explicitly considers the properties of environment dynamics and models the dynamics in a decomposing manner. \alg contains two components: SD2 and D2P. SD2 decomposes the environment dynamics into several sub-dynamics according to the dynamics-action relation. D2P constructs a decomposing prediction model according to the result of SD2. When we combine \alg, the performance of existing MBRL algorithms is significantly improved. Currently, this work only considers the decomposition at the dimension level, and for future work, it is worthwhile investigating how to decompose environment dynamics at the object level, which can further improve the interpretability and generalizability of \alg.

% In this paper, we propose a novel world model construction framework that leverages environment dynamics properties: Environment Dynamics Decomposition (ED2). Our \alg explicitly decomposes the environment dynamics into several sub-dynamics based on the analysis of correlations between action dimensions and state dimensions. Each sub-dynamics is predicted separately and then combined into a full dynamics prediction. In this way, \alg establishes a more accurate world model, which greatly improves the asymptotic performance of existing MBRL algorithms. Currently, this work consider the decomposition on dimension levels, and for future work, it is worthwhile investigating how to decompose environment dynamics at the object level, which can further improve the interpretability and generalizability of \alg.
%%%%%%%%%%%%%%%%%%%%%%%%%%%%%%%%%%%%%%%%%%%%%%%%%%%%%%%%%%%%

\nocite{*}
\bibliographystyle{IEEEtran}
\bibliography{ed2}

% \newpage

 

% \end{IEEEbiography}



\vfill

\end{document}